\pdfoutput=1
\documentclass{article}
\usepackage{iclr2027_conference,times}
\usepackage{amsmath,amssymb}
\usepackage{graphicx,booktabs,tabularx,xcolor}
\usepackage{microtype}
\usepackage{hyperref,url}
\usepackage[percent]{overpic}
\hypersetup{hidelinks,pdfauthor={Botong Zhao, Fang Yu, Tim Yu, Senhua Zhu, Xinyuan Chen, Yue Lu},pdftitle={PAVE: Separates Prediction from Preference in World-Action Policies}}
\newcommand{\safeimage}[2][]{\includegraphics[#1]{#2}}
\makeatletter
\let\pavecaptionwithcolon\@makecaption
\long\def\pavecaptionnocolon#1#2{%
  \vskip\abovecaptionskip
  \sbox\@tempboxa{#1\ #2}%
  \ifdim \wd\@tempboxa >\hsize
    #1\ #2\par
  \else
    \global\@minipagefalse
    \hb@xt@\hsize{\hfil\box\@tempboxa\hfil}%
  \fi
  \vskip\belowcaptionskip}
\newcommand{\PAVECaptionsNoColon}{\let\@makecaption\pavecaptionnocolon}
\newcommand{\PAVECaptionsWithColon}{\let\@makecaption\pavecaptionwithcolon}
\makeatother

\title{PAVE: Separates Prediction from\\Preference in World-Action Policies}
\author{%
\normalfont\parbox{\dimexpr\textwidth-2\tabcolsep\relax}{\centering
Botong Zhao$^{1,2,3}$\quad Fang Yu$^{1,3}$\quad Tim Yu$^{2}$\\[2pt]
Senhua Zhu$^{2}$\quad Xinyuan Chen$^{3}$\quad Yue Lu$^{1,*}$\\[6pt]
{\small
$^{1}$Multi-Dimensional Information Processing Laboratory,\\
East China Normal University\\[2pt]
$^{2}$Lab for Brain-Inspired Embodied Intelligence,\\
EBKernel Technologies Co., Ltd\\[2pt]
$^{3}$Shanghai Artificial Intelligence Laboratory\\[4pt]
$^{*}$Corresponding author}
}}
\iclrfinalcopy
\begin{document}
\maketitle
\lhead{Preprint}
\PAVECaptionsNoColon
\begin{abstract}
\begingroup
An unsuccessful robot rollout records valid transitions alongside behavior a policy should improve. PAVE (Predictive Alignment and Value-Guided Evolution) equips world-action policies with trajectory-adaptive predictive alignment and action-conditioned value learning. Local and trajectory-relative joint-embedding targets train shared actor representations across temporal scales, retaining valid transitions from both successes and failures. A dual-head critic couples future-latent prediction with return estimation; episode-excluded values assign relative-quality conditions to action chunks. Both quality conditions train the actor, and deployment selects positive. Matched-data actor comparisons improve on local prediction with quality conditioning by $5.0/4.5$ percentage points on LIBERO-Plus/RoboTwin Random. A matched critic ablation isolates the benefit of future-latent learning. On S1, PAVE achieves $85.7\pm3.5\%/92.0\pm2.0\%$ on organization/pouring versus $80.0\pm1.0\%/84.7\pm1.5\%$ for the matched control (mean $\pm$ sample SD over three training seeds; 100 trials per seed and task). Target encoding, predictive heads, and critics remain offline.

\endgroup
\end{abstract}

\section{Introduction}
\label{sec:introduction}

A robot can fail a task without invalidating what its sensors recorded. For a world-action policy trained to predict and act through shared representations, this creates a \emph{supervision-role mismatch}. Observed futures describe what happened, whereas action learning must distinguish what is worth repeating. We study how to retain the former without indiscriminately imitating the latter during policy evolution.

Generalist vision-language-action (VLA) policies connect perception and instructions to robot control, drawing on web knowledge, robot demonstrations, and direct generative action learning \citep{zitkovich2023rt2,ghosh2024octo,kim2024openvla,black2024pi0,physicalintelligence2025pi05,chi2023diffusionpolicy}. Experience-driven post-training adds successful, unsuccessful, and recovery behavior \citep{physicalintelligence2025pistar06,wu2026flowpro,yan2026redflow,yang2026aloe}. The question is not only which experience to retain, but \emph{what the same experience should supervise} when predictive and behavioral objectives share an actor.

Consider an object dropped after a successful approach. The recorded motion remains valid predictive evidence, while the executed chunks can differ in behavioral quality. A common success filter discards useful transitions; unconditioned imitation of the whole pool ignores preference. Neither is required. Valid transitions can supervise prediction while their actions train a quality-conditioned distribution. Validity concerns data integrity, not task success. Corruption or unknown-outcome interruptions still require masking or segmentation; an unsuccessful episode does not automatically make every action negative.

A second challenge is adapting predictive supervision to the task's changing temporal scale. Future-oriented objectives, foresight, histories, and progress annotations provide temporal structure \citep{miao2026jepavla,sun2026vlajepa,yuan2026fastwam,li2026wav,wang2026oflow,ma2026stpi,wang2025lola,mao2026arm}. A fixed offset covers a changing fraction of the remaining episode and clips at its endpoint. PAVE introduces trajectory-relative predictive alignment: horizon-conditioned predictors align shared actor states with joint current--future embeddings at normalized remaining-trajectory positions. Alongside a local objective, these targets span broad changes early and finer changes near termination, adapting the supervision scale throughout each recorded rollout.

\begin{figure}[htbp]
    \centering
    \includegraphics[width=\textwidth]{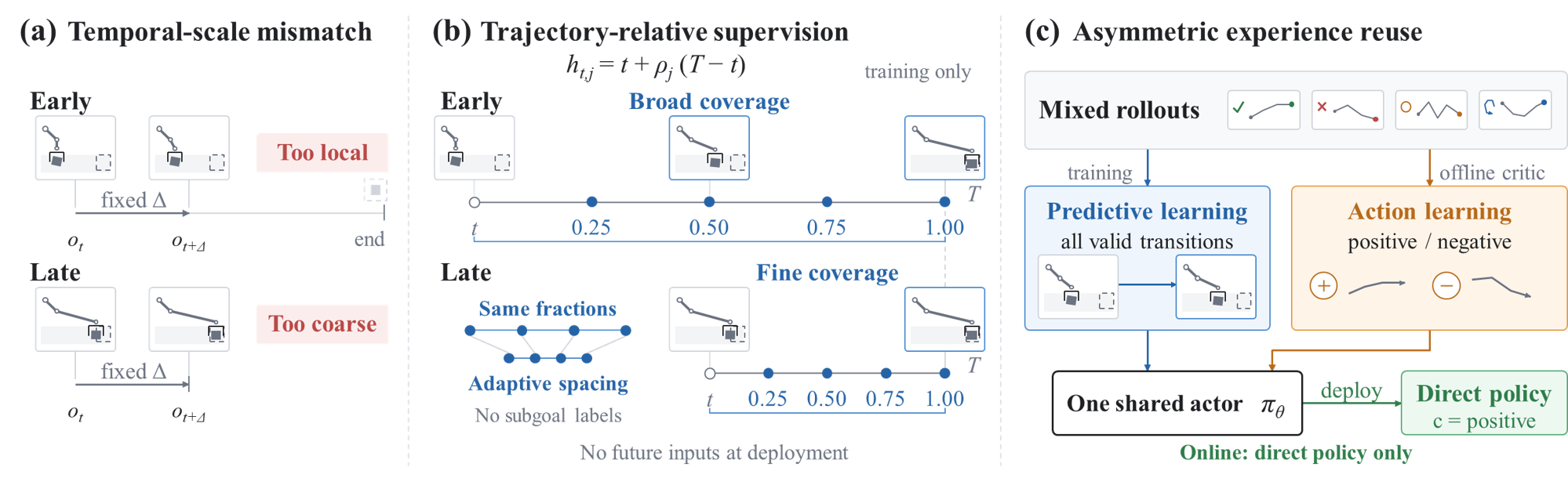}
    \caption{\textbf{The same experience, different supervision roles.} (a) Fixed offsets can mismatch the remaining temporal scale. (b) Trajectory-relative anchors provide broad-to-local coverage of recorded changes, alongside local targets. (c) All valid transitions support prediction; relative-quality conditions govern action supervision. Both positive and negative action examples remain in one shared actor. The asymmetry concerns supervision roles, not separate parameters or a failure-to-negative rule.}
    \label{fig:intro_concept}
\end{figure}

\textbf{PAVE} (Figure~\ref{fig:intro_concept}) couples this temporal alignment with predictive value learning. The actor learns across local and trajectory-relative horizons, while the critic learns action consequences and returns through shared state features. Episode-excluded values then condition action learning on relative chunk quality. This coupling extracts transition structure from mixed-quality experience and directs behavioral updates toward better chunks. Predictive losses update shared representations while action generation uses current observations, instructions, and the selected quality condition.

Our contributions are three coupled learning mechanisms: trajectory-adaptive multi-scale alignment for the actor, an action-conditioned predictive critic that learns future latents jointly with returns, and episode-excluded quality conditioning with asymmetric reuse of mixed-quality experience. Together they connect temporal representation learning, chunk evaluation, and policy evolution. Matched actor controls yield $5.0/4.5$-point gains on LIBERO-Plus/RoboTwin Random; a critic control isolates future-latent supervision, and S1 evaluates the resulting policy evolution on physical hardware. All predictive and value-learning modules are removed from the online control path.

\section{Related Work}
\begingroup
\paragraph{Generalist policies and predictive representations.}
RT-2 and Octo establish complementary foundations through web-knowledge transfer and flexible robot pretraining \citep{zitkovich2023rt2,ghosh2024octo}. PAVE studies experience reuse within a shared actor rather than replacing those pretraining paradigms. JEPA-VLA, VLA-JEPA, Fast-WAM, Being-H0.7, StageWAM, and JEPA-WAM explore future-oriented representations \citep{miao2026jepavla,sun2026vlajepa,yuan2026fastwam,luo2026beingh07,liu2026stagewam,lin2026jepawam}. PAVE combines joint current--future targets with a direct actor and studies temporal target placement, mixed-quality supervision, and action-conditioned predictive value learning.

\paragraph{Foresight tokens and planning auxiliaries.}
WAV and OFlow use latent foresight during action generation \citep{li2026wav,wang2026oflow}. InternVLA-A1.5 instead supervises contextual foresight tokens through a frozen video generator; its video loss updates those tokens, which condition actions, while the generator is discarded at deployment \citep{ma2026internvla15}. PHR-VLA aligns action-token representations with future latent dynamics through a training-only head \citep{soleymanzadeh2026phrvla}. PAVE likewise removes its auxiliary heads, but their losses supervise visual-position states, not its action-only query parameters. It uses joint current--future targets at remaining-episode fractions and separately annotates behavior quality. These predictive interfaces could complement quality conditioning, but their combination is untested here. Without matched comparisons to InternVLA-A1.5 or PHR-VLA, architectural differences establish neither performance nor total-training-cost superiority.

\paragraph{Quality conditioning and action multimodality.}
RECAP uses advantage conditioning; FlowPRO uses intervention-and-rollback preferences; RedFlow derives corrections from failures; ALOE estimates action-chunk values \citep{physicalintelligence2025pistar06,wu2026flowpro,yan2026redflow,yang2026aloe}. VFP models action multimodality with a variational prior, optimal-transport alignment, and a mixture-of-experts flow decoder \citep{zhai2025vfp}. Multiple modes can all be desirable, whereas PAVE's label expresses a relative preference among recorded chunks. PAVE retains a direct flow decoder and studies how preference interacts with continued predictive supervision. Our comparisons distinguish method-level adaptations from full-system reproductions and report paired FlowPRO-style supervision separately.

\paragraph{Temporal structure and progress.}
ST-$\pi$, LoLA, and ARM use structured prompts, long histories, and progress annotations, respectively \citep{ma2026stpi,wang2025lola,mao2026arm}. PAVE indexes recorded temporal positions rather than semantic stages, retaining valid failed endpoints. Encoder comparisons with V-JEPA~2.1, DINOv2, and LingBot-Vision hold the target encoder and critic fixed \citep{murlabadia2026vjepa21,oquab2023dinov2,fu2026lingbotvision}.
\endgroup

\section{Proposed Method}
\label{sec:method}
Let $\mathcal D_{\rm demo}$ denote expert demonstrations and $\mathcal D_k$ the deployment trajectories collected in round $k$, with $\mathcal D_{\leq k}=\bigcup_{j=1}^{k}\mathcal D_j$. PAVE learns
$\pi_\theta(a_{t,\ldots,t+H-1}\mid o_t,l,q_t,c_t)$,
where $o_t$ contains the available camera views, $l$ is the task instruction, $q_t$ is proprioception, and $c_t\in\{\mathrm{positive},\mathrm{negative},\mathrm{null}\}$ is a behavior-quality condition. The trajectory outcomes are used for offline value supervision, not supplied as future information to the deployed actor.

\begin{figure}[htbp]
\centering
\begin{overpic}[width=\textwidth]{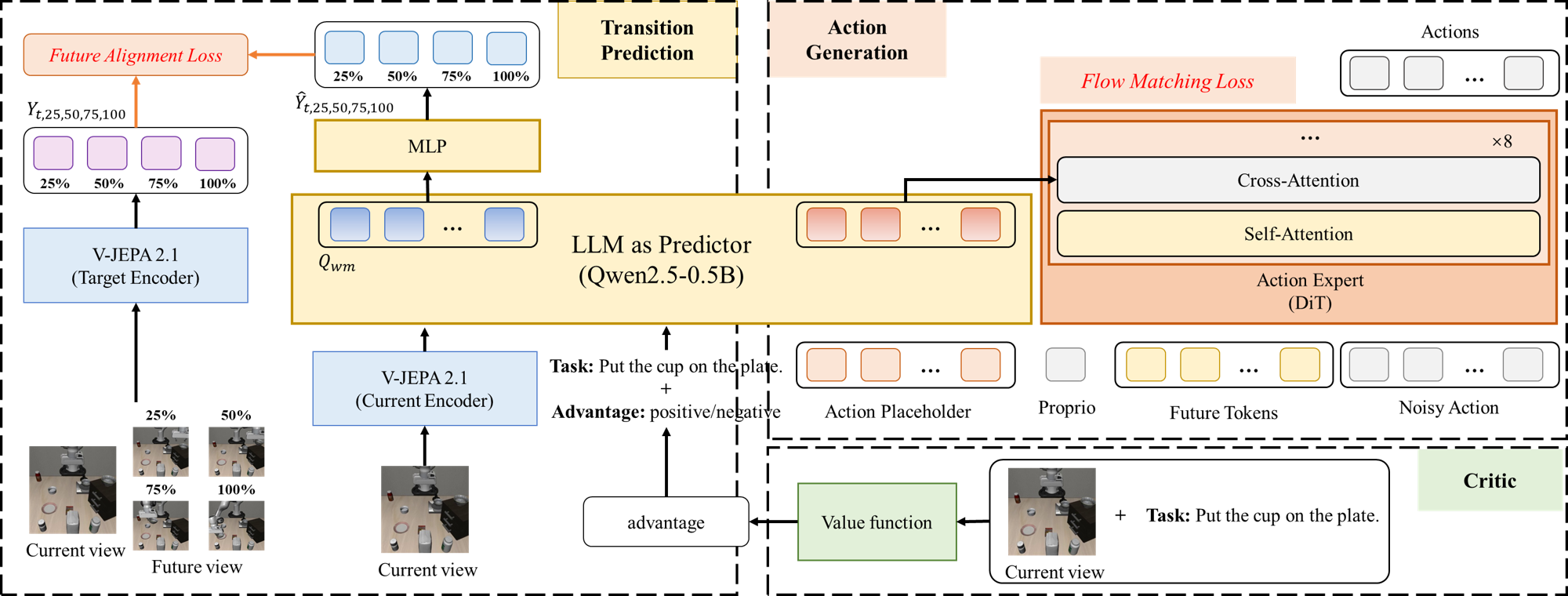}
\put(94.0,28.25){\color[rgb]{.9843,.8980,.8353}\rule{.032\textwidth}{.019\textwidth}}
\put(94.2,28.85){\makebox(0,0)[l]{\fontsize{5.1}{5.5}\selectfont\texttimes16}}
\put(73.0,9.8){\color{white}\rule{.115\textwidth}{.022\textwidth}}
\put(78.75,11.55){\makebox(0,0){\fontsize{4.9}{5.1}\selectfont Learnable queries}}
\end{overpic}
\caption{PAVE separates supervision roles within a shared actor. The predictive branch aligns visual-position states with frozen current--future targets; the action branch uses quality-conditioned states. Relative targets are shown; local prediction is retained. The critic produces offline chunk labels, not online control. The $16$-layer decoder retains learnable queries. The actor predictive MLP omits the recorded action chunk to avoid a target-action leakage path in flow matching and to learn the expected future from available observations.}
\label{fig:method_overview}
\end{figure}
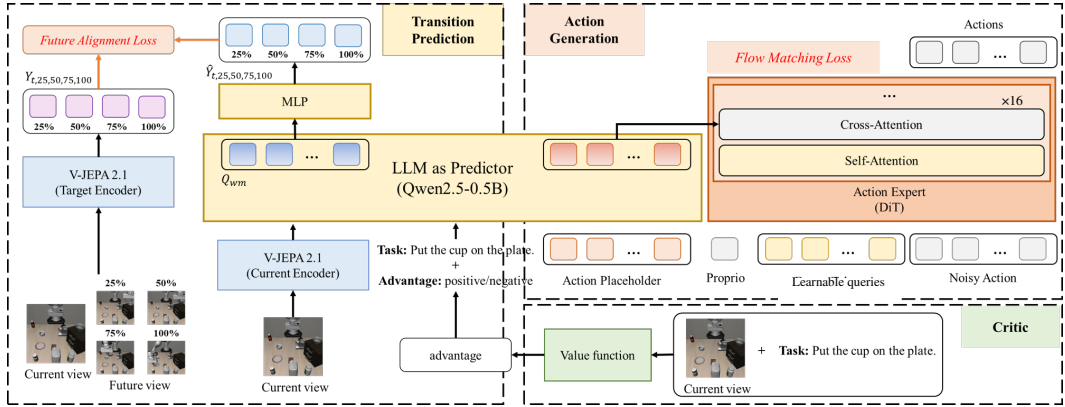

\begingroup
\paragraph{Different supervision roles, shared parameters.}
As shown in Figure~\ref{fig:method_overview}, PAVE assigns validity masks to predictive pairs and relative-quality conditions to executed actions; one scalar success label does not govern both objectives. The actor uses a frozen visual encoder/projector, a Qwen backbone adapted with low-rank adapters (LoRA), and a trainable action decoder. Both losses update the Qwen adapters. The separate critic jointly trains a fusion MLP and independent return and predictive heads. Deployment removes target encoding, auxiliary heads, and critic evaluation (Appendix~\ref{app:gradientledger}).
\endgroup

\subsection{Predictive Alignment in Trajectory-Relative Time}
\label{sec:mh}
\paragraph{Direct actor and learnable query tokens.}
Frozen V-JEPA~2.1 features enter a causal Qwen backbone through a frozen projector; pretrained Qwen weights are adapted only through LoRA \citep{murlabadia2026vjepa21,yang2024qwen25,hu2022lora}. Visual tokens precede instruction, quality, and 64 action placeholders:
\begin{equation}
 (V_t,M_t)=Q_\theta([w_0;P(E(o_t));w_{1,\ldots,L_w};c_t;q^a_{1,\ldots,64}]),
 \label{eq:actorstates}
\end{equation}
where $V_t,M_t$ select visual- and action-position states. The decoder also receives proprioception $q_t$ and $K_q=32$ learnable future-query tokens $F^{\rm learn}$ \citep{lin2026jepawam}. These slots receive only action-loss gradients; predictive semantics are unestablished (Appendix~\ref{app:querydiagnostics}). Predictive targets and outputs never enter the decoder. Appendix~\ref{app:actorimpl} specifies token meanings and dimensions; Appendix Table~\ref{tab:hyperparams} gives defaults.

For actions $a\in\mathbb R^{H\times d_a}$, Gaussian noise $\epsilon$, and flow time $\tau$, let $x_\tau=(1-\tau)\epsilon+\tau a$ and $u^*=a-\epsilon$. The velocity decoder minimizes
\begin{equation}
 \mathcal L_{\rm FM}=\frac{\sum_{b,h,d}m^a_{bhd}
 [u_\theta(x_\tau,\tau,M_t,q_t,F^{\rm learn})-(a-\epsilon)]_{bhd}^{2}}
 {\sum_{b,h,d}m^a_{bhd}},
 \label{eq:fm}
\end{equation}
masking invalid dimensions, padding, and inconsistent chunks. LIBERO uses two views and seven action dimensions; bimanual settings use their own $n_v,d_a$. Controlled actors use $H=20$; published JEPA-WAM horizons differ (Appendix~\ref{app:protocol}).

\paragraph{Joint current--future targets.}
The frozen target encoder jointly encodes each camera's current and recorded future images as a two-frame clip:
$E_{\rm pair}(o_t,o_h)=\operatorname{Concat}_{v=1}^{n_v}
 E_{\rm tgt}(\operatorname{Stack}_{\rm time}(I_t^v,I_h^v))
 \in\mathbb R^{N_v\times1024}$.
Predictors map visual-position states to joint embeddings, preserving camera and patch positions. The Figure~\ref{fig:method_overview} predictive MLP omits the recorded action chunk: passing ground-truth future actions through shared conditioning would risk leaking the flow-matching target. Its purpose is the expected future given available observations, rather than simulation of a specified action sequence.

\paragraph{Local targets and normalized remaining time.}
For last observation index $T$, define $h_t^\delta=\min(t+\delta,T)$ and $Y_t^\delta=\operatorname{sg}[E_{\rm pair}(o_t,o_{h_t^\delta})]$, with local offset $\delta=31$. Predictor $P_\delta$ minimizes
\begin{equation}
 \mathcal L_{\rm local}=\frac{\sum_t m_t^\delta d_{\cos}(P_\delta(V_t),Y_t^\delta)}{\sum_t m_t^\delta},
 \label{eq:local}
\end{equation}
where $d_{\cos}$ averages $1-\cos$ over corresponding patch tokens. Additional targets occupy fixed fractions $\rho=\{0.25,0.50,0.75,1.00\}$ of the remaining recorded trajectory:
\begin{align}
 h_{t,j}&=\operatorname{clip}(\operatorname{round}[t+\rho_j(T-t)],t,T),\label{eq:horizonindex}\\
 Y_{t,j}&=\operatorname{sg}[E_{\rm pair}(o_t,o_{h_{t,j}})],\qquad
 \widehat Y_{t,j}=P_{\rm MH}(V_t+e_j),\\
 \mathcal L_{\rm MH}&=\frac{\sum_{t,j}m_{t,j}d_{\cos}(\widehat Y_{t,j},Y_{t,j})}{\sum_{t,j}m_{t,j}}.
 \label{eq:mh}
\end{align}
Before rounding, $(h_{t,j}-t)/(T-t)=\rho_j$ for $t<T$: each horizon preserves its relative position as absolute offsets contract along the rollout. Horizon embeddings $e_j\in\mathbb R^{896}$ broadcast over visual positions. Masks exclude episode, policy-version, intervention/source, and unknown-outcome crossings; valid failed endpoints remain observed targets. Future observations and $T$ are training-only (degeneracy checks: Appendix~\ref{app:clarifications}).

\paragraph{Shared optimization, different supervision roles.}
Let $\theta_{\rm sh},\theta_{\rm act},\theta_{\rm pred}$ denote shared-backbone, action-path, and auxiliary-prediction parameters. The actor objective is
\begin{equation}
\begin{aligned}
 \mathcal L_{\rm actor}={}&\mathcal L_{\rm FM}(\theta_{\rm sh},\theta_{\rm act};c_t)
 +\lambda_{\rm local}\mathcal L_{\rm local}(\theta_{\rm sh},\theta_{\rm pred};m^\delta) +\lambda_{\rm MH}\mathcal L_{\rm MH}(\theta_{\rm sh},\theta_{\rm pred};m),
\end{aligned}
 \label{eq:actorloss}
\end{equation}
with $\lambda_{\rm local}=\lambda_{\rm MH}=0.5$ and the masks above. Prediction retains valid pairs across outcomes; $c_t$ conditions action quality. Negative examples remain in training and can influence shared parameters.

\paragraph{Gradient paths and the future-information boundary.}
Both objectives update Qwen adapters. Only action loss updates the decoder and learnable queries; only predictive losses update auxiliary heads and horizon embeddings. Actor gradients never update the critic (Appendix~\ref{app:actorimpl}). Stop-gradient targets are encoded separately, outside actor inputs and attention; $Y_{t,j},\widehat Y_{t,j},h_{t,j},T$ never enter the decoder. Prediction shapes representations through parameter updates alone.

Causal $V_t$ cannot attend to later task/quality tokens, whereas $M_t$ can. Thus prediction averages over unresolved futures rather than encoding instruction-specific plans. It can improve shared Qwen adapters while action supervision retains task specificity; the visual encoder stays frozen. Fine-detail preservation is not guaranteed (Appendix~\ref{app:clarifications}).

\subsection{Action-Conditioned Predictive Value Learning}
\label{sec:value}
\begin{figure}[htbp]
\centering
\includegraphics[width=\textwidth]{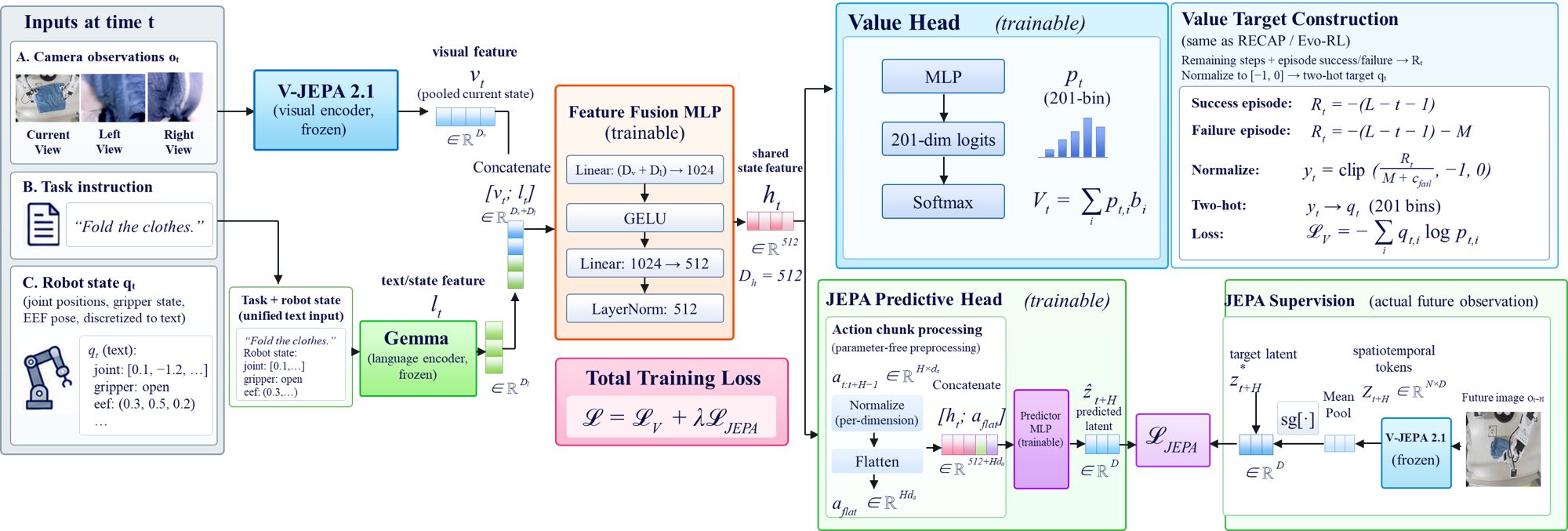}
\caption{Predictive value learning. Frozen V-JEPA~2.1 and Gemma features feed a shared fusion MLP. The return head estimates state value; a separate JEPA head predicts the actual future from the shared feature and executed chunk. Both losses update fusion. Actions enter only the predictive head, and future images only its frozen target branch. Offline annotation runs the value path.}
\label{fig:value_function}
\end{figure}

We design a dual-head critic for offline behavioral annotation (Figure~\ref{fig:value_function}). Return estimation and action-conditioned future prediction train a shared state representation, aligning the auxiliary prediction span with the chunks being scored. The critic is separate from the actor; its fixed labels condition behavior without filtering predictive evidence.

\paragraph{Shared state, separate heads.}
Frozen V-JEPA~2.1 pools current/short-history observations into $v_t$; frozen Gemma pools task text and discretized proprioception into $g_t$. Their concatenation feeds a $(D_v+D_l)\!\to\!1024\!\to\!512$ fusion MLP with GELU and final LayerNorm, producing $h_t^c$. A return MLP yields $p_\phi(b_i\mid s_t)$ on $b_i=-1+i/200$, $i=0,\ldots,200$, for $s_t=(o_{\leq t},l,q_t)$. A separate predictive MLP receives $h_t^c$ and the normalized, flattened executed chunk $A_t=a_{t:t+H-1}$:
\begin{equation}
 \widehat z_{t+H}=P_\psi([h_t^c;\operatorname{Flatten}(\operatorname{Norm}(A_t))]),\quad
 z^*_{t+H}=\operatorname{sg}[\operatorname{Pool}(E_v(o_{\leq t+H}))].
 \label{eq:criticprediction}
\end{equation}
Both latents have width $D_v$; there is no separate action encoder or learned target projection. Value sees only current/history information. Unlike the actor MLP in Figure~\ref{fig:method_overview}, the offline auxiliary uses executed actions to learn their consequences.

\paragraph{Return and predictive supervision.}
For $L\leq M_{\rm task}$ valid actions, nonterminal rewards are $-1$ and the final reward is $0$ on success or $-M_{\rm task}$ on valid failure. With $C=2M_{\rm task}$,
\begin{equation}
 y_t=\operatorname{clip}\!\left(C^{-1}\!\sum_{i=t}^{L-1}r_i,-1,0\right),\quad
 \mathcal L_{\rm value}=-\sum_{i=0}^{200}p^*_{t,i}\log p_\phi(b_i\mid s_t),\quad
 V_\phi(s)=\sum_i b_i p_\phi(b_i\mid s),
 \label{eq:value}
\end{equation}
where $p_t^*$ is the two-hot interpolation of $y_t$ \citep{bellemare2017distributional}. The critic objective is
\begin{equation}
 \mathcal L_{\rm critic}=\mathcal L_{\rm value}+\lambda_{\rm J}\mathcal L_{\rm J},\qquad
 \mathcal L_{\rm J}=\mathbb E_{\rm valid}[1-\cos(\widehat z_{t+H},z^*_{t+H})].
 \label{eq:criticloss}
\end{equation}
Both losses update fusion; each head receives its own loss. Complete valid $H$-step transitions support $\mathcal L_{\rm J}$ regardless of success; unknown-outcome censoring receives no invented return. Latent similarity is an auxiliary constraint, not a reward or value label. Appendix~\ref{app:critic} gives masks, gradient paths, and parameter accounting.

\paragraph{Episode-excluded chunk labels.}
Partition deployment episodes into $F$ folds \citep{chernozhukov2016crossfit}. Critic $\phi_k^{(-f(e))}$ excludes scored episode $e$ from both return and predictive fitting, preprocessing, and model selection. Both endpoint values use this same critic. With $m_t=\min(N,L-t)$ and $\bar r_t=r_t/C$,
\begin{equation}
 A_{t,e}^{(N),\rm OOF}=\sum_{j=0}^{m_t-1}\bar r_{t+j}
 +\mathbf1[t+N<L]V_{\phi_k^{(-f(e))}}(s_{t+m_t})
 -V_{\phi_k^{(-f(e))}}(s_t).
 \label{eq:advantage}
\end{equation}
True terminal chunks do not bootstrap. With $N=H=20$, a full nonterminal chunk has score $V(s_{t+H})-V(s_t)-H/C$. Within each task, the top $30\%$ of eligible complete chunks are positive and the rest negative, with fixed tie handling and separate expert-correction rules. Scores compare observed rewards and reached states with a behavior-mixture baseline; positive need not mean $A>0$ or success. Cross-fitting prevents episode self-fitting, not bias or poor support. Binary labels discard within-class distinctions; annotation uses only the value path, and neither critic head is deployed.

\subsection{Asymmetric Experience Reuse for Policy Evolution}
\label{sec:evolution}
Asymmetric reuse means that a valid transition remains predictive evidence even when its action has a negative quality condition. Negative-conditioned actions also remain training examples. Instructions append a positive or negative advantage condition. Omitting the condition gives null, with condition dropout $0.3$. Actor training treats labels as fixed. Demonstrations are positive; rollout chunks use cross-fitted labels. Critics fit cumulative deployment experience, while actor replay retains demonstrations. Appendix tables abbreviate this conditioning as ACP.

Each round jointly fits fold-excluded critic fusion and heads, labels $\mathcal D_{\leq k}$, and trains on $\mathcal D_{\rm train}^k=\mathcal D_{\rm demo}\cup\widetilde{\mathcal D}_{\leq k}$ while sampling demonstrations three times as often as evolution data. Common-base initialization controls parameter-path dependence. The joint training schedule uses three rounds, $F=2$, $30{,}000$ actor updates per round, and $8{,}000$ per fold-specific critic; collection budgets differ between simulation and S1 (Appendix Table~\ref{tab:hyperparams}).

At deployment, $c_t=\mathrm{positive}$ and a $16$-layer DiT generates chunks in four Euler steps. ``Direct'' excludes online target encoding, prediction heads, critic calls, rollout search, and MPC; it still requires four decoder evaluations. Learnable future-query tokens remain in the actor.

\PAVECaptionsWithColon
\section{Experiments}
\label{sec:experiments}
{We evaluate PAVE on standard manipulation, held-out perturbations, controlled experience reuse, and physical S1 tasks. Published results provide context; demonstration-only controls isolate predictive supervision; same-pool post-training controls compare PAVE with conditional extraction and unlabeled behavior cloning (BC). We relate differences to the factor varied by each control. Table~\ref{tab:rounds} reports the evolution ablation with the JEPA-assisted critic in Section~\ref{sec:value}.}

\subsection{Experimental Setup and Comparison Scope}
\paragraph{Benchmark tracks.}
LIBERO covers spatial, object, goal, and long-horizon manipulation; LIBERO-Plus introduces controlled perturbations \citep{liu2023libero,fei2025liberoplus}. The source-aligned generalization track trains on LIBERO data and permits evolution on disjoint clean-training initializations. LIBERO-Plus evaluation episodes, images, outcomes, and initial-state manifests are excluded from actor fitting, critic fitting, and hyperparameter selection. LIBERO-Plus thus measures generalization from clean-training experience, not adaptation to test perturbations.

For RoboTwin~2.0 \citep{chen2025robotwin}, JEPA-WAM reports $20$ Clean-trained tasks in two groups, evaluated under both Clean and Random. Our runs use these groups and their evaluation manifests. Random denotes joint randomization over this $20$-task track, distinct from a $50$-task study or an average of isolated one-factor perturbations.

\paragraph{Literature anchors versus matched controls.}
{Published entries reproduce the named JEPA-WAM tables, including their results for other methods, without adding unreported standard deviations. Compact JEPA-WAM and pretrained $\pi_{0.5}$ are different systems. Our demonstration-only \emph{JEPA-WAM} control shares PAVE's actor interface and clean demonstrations. The \emph{local-prediction + quality-conditioning control} also matches PAVE's cumulative rollout pool, initialization, and actor-update budget, but omits trajectory-relative targets. Appendix Tables~\ref{tab:interaction} and~\ref{tab:routing} compare training and data-use rules on this fixed pool. Simulation collection adds $96$ episodes per training group per seed per round, totaling $288$ over three rounds. Published clean-only policies provide reference context, not equal-data post-training comparisons.}

\paragraph{Implementation and reporting.}
Unless otherwise stated, controlled runs use $H=N=20$, four flow integration steps, and episode-disjoint two-fold cross-fitting ($F=2$). Local simulation results report means and sample standard deviations across five independent training seeds. Appendix~\ref{app:configuration} (Appendix Table~\ref{tab:hyperparams}) summarizes actor, critic, and policy-evolution settings; Appendix~\ref{app:critic} details the dual-head critic.

\subsection{Policy Performance under Matched Experience}
\begin{table}[htbp]
\centering\scriptsize
\caption{LIBERO success (\%). Published rows reproduce JEPA-WAM Table~1 \citep{lin2026jepawam}; local rows use the matched protocol described above. PT denotes large-scale robot-policy pretraining.}
\label{tab:libero}
\begin{tabular}{@{}lcccccc@{}}\toprule
Method & PT & Spatial & Object & Goal & Long & Avg. \\\midrule
\multicolumn{7}{@{}l}{\textit{{Published reference systems}}} \\
Diffusion Policy (published) & No & 78.3 & 92.5 & 68.3 & 50.5 & 72.4 \\
Fast-WAM (published) & No & 98.2 & 100.0 & 97.0 & 95.2 & 97.6 \\
JEPA-WAM (published) & No & 95.6 & 99.4 & 97.2 & 94.6 & 96.7 \\
$\pi_{0.5}$ (published) & Yes & 98.6 & 98.2 & 98.4 & 92.4 & 96.9 \\
VLA-JEPA (published) & Yes & 96.2 & 99.6 & 97.2 & 95.8 & 97.2 \\
$\pi_{0.5}$+JEPA (published) & Yes & 99.0 & 98.0 & 97.6 & 96.4 & 97.8 \\\midrule
\multicolumn{7}{@{}l}{\textit{{Controlled local policies}}} \\
JEPA-WAM (demo-only) & No & \textcolor{black}{\ensuremath{95.5\,\pm\,0.5}} & \textcolor{black}{\ensuremath{99.3\,\pm\,0.2}} & \textcolor{black}{\ensuremath{97.1\,\pm\,0.3}} & \textcolor{black}{\ensuremath{94.5\,\pm\,0.5}} & \textcolor{black}{\ensuremath{96.6\,\pm\,0.4}} \\
{Local prediction + quality conditioning} & No & \textcolor{black}{\ensuremath{97.0\,\pm\,0.6}} & \textcolor{black}{\ensuremath{99.4\,\pm\,0.3}} & \textcolor{black}{\ensuremath{97.4\,\pm\,0.5}} & \textcolor{black}{\ensuremath{96.2\,\pm\,0.3}} & \textcolor{black}{\ensuremath{97.5\,\pm\,0.3}} \\
PAVE (ours) & No & \textcolor{black}{\ensuremath{98.6\,\pm\,0.4}} & \textcolor{black}{\ensuremath{99.5\,\pm\,0.3}} & \textcolor{black}{\ensuremath{98.9\,\pm\,0.3}} & \textcolor{black}{\ensuremath{98.5\,\pm\,0.5}} & \textcolor{black}{\ensuremath{98.9\,\pm\,0.2}} \\\bottomrule
\end{tabular}

\end{table}

\begin{table}[htbp]
\centering\scriptsize
\caption{LIBERO-Plus success (\%). Published rows reproduce JEPA-WAM Table~2 \citep{lin2026jepawam}. Controlled actors share clean demonstrations and evaluation manifests; evolution methods additionally share rollouts, initialization, and updates. No LIBERO-Plus evaluation experience enters training.}
\label{tab:liberoplus}
\resizebox{\textwidth}{!}{\begin{tabular}{@{}lccccccccc@{}}\toprule
Method & PT & Camera & Robot & Language & Light & Backgr. & Noise & Layout & Avg. \\\midrule
\multicolumn{10}{@{}l}{\textit{{Published reference systems}}} \\
JEPA-WAM (published) & No & 79.2 & 59.2 & 68.2 & 93.3 & 94.6 & 83.6 & 76.1 & 79.2 \\
VLA-JEPA (published) & Yes & 63.3 & 67.1 & 85.4 & 95.6 & 93.6 & 66.3 & 85.1 & 79.5 \\
$\pi_{0.5}$ (published) & Yes & 69.4 & 75.3 & 82.6 & 96.7 & 96.8 & 84.3 & 86.2 & 84.5 \\
$\pi_{0.5}$+JEPA (published) & Yes & 66.0 & 82.0 & 86.5 & 96.8 & 96.0 & 88.3 & 88.3 & 86.3 \\\midrule
\multicolumn{10}{@{}l}{\textit{{Controlled local policies}}} \\
JEPA-WAM (demo-only) & No & \textcolor{black}{\ensuremath{78.8\,\pm\,1.0}} & \textcolor{black}{\ensuremath{59.4\,\pm\,1.1}} & \textcolor{black}{\ensuremath{67.9\,\pm\,0.8}} & \textcolor{black}{\ensuremath{93.2\,\pm\,0.5}} & \textcolor{black}{\ensuremath{94.2\,\pm\,0.4}} & \textcolor{black}{\ensuremath{83.4\,\pm\,0.8}} & \textcolor{black}{\ensuremath{76.1\,\pm\,0.9}} & \textcolor{black}{\ensuremath{79.0\,\pm\,0.8}} \\
{Local prediction + quality conditioning} & No & \textcolor{black}{\ensuremath{81.0\,\pm\,0.9}} & \textcolor{black}{\ensuremath{64.0\,\pm\,1.0}} & \textcolor{black}{\ensuremath{72.5\,\pm\,0.6}} & \textcolor{black}{\ensuremath{94.3\,\pm\,0.5}} & \textcolor{black}{\ensuremath{95.2\,\pm\,0.4}} & \textcolor{black}{\ensuremath{87.0\,\pm\,0.5}} & \textcolor{black}{\ensuremath{80.0\,\pm\,1.0}} & \textcolor{black}{\ensuremath{82.0\,\pm\,0.7}} \\
PAVE (ours) & No & \textcolor{black}{\ensuremath{86.0\,\pm\,0.6}} & \textcolor{black}{\ensuremath{71.0\,\pm\,1.0}} & \textcolor{black}{\ensuremath{80.5\,\pm\,0.9}} & \textcolor{black}{\ensuremath{97.4\,\pm\,0.4}} & \textcolor{black}{\ensuremath{97.4\,\pm\,0.3}} & \textcolor{black}{\ensuremath{91.2\,\pm\,0.9}} & \textcolor{black}{\ensuremath{85.7\,\pm\,0.7}} & \textcolor{black}{\ensuremath{87.0\,\pm\,0.7}} \\\bottomrule
\end{tabular}}

\end{table}

\begin{table}[htbp]
\centering\scriptsize
\caption{RoboTwin~2.0 success (\%) on JEPA-WAM's 20-task manifest \citep{lin2026jepawam}; Random applies joint randomization. Published rows give source-protocol context; local rows follow the matched demonstration-only/evolution protocols described above.}
\label{tab:robotwin}
\begin{tabular}{@{}llcc@{}}\toprule
Method & PT & Clean & Random \\\midrule
\multicolumn{4}{@{}l}{\textit{{Published reference systems}}} \\
Diffusion Policy (published) & No & 48.0 & 1.6 \\
JEPA-WAM (published) & No & 79.9 & 36.9 \\
$\pi_{0.5}$ (published) & Yes & 75.4 & 37.2 \\
$\pi_{0.5}$+JEPA (published) & Yes & 84.6 & 37.5 \\\midrule
\multicolumn{4}{@{}l}{\textit{{Controlled local policies}}} \\
JEPA-WAM (demo-only) & No & \textcolor{black}{\ensuremath{79.6\,\pm\,1.1}} & \textcolor{black}{\ensuremath{36.7\,\pm\,0.8}} \\
{Local prediction + quality conditioning} & No & \textcolor{black}{\ensuremath{82.5\,\pm\,0.9}} & \textcolor{black}{\ensuremath{40.7\,\pm\,1.4}} \\
PAVE (ours) & No & \textcolor{black}{\ensuremath{86.8\,\pm\,1.3}} & \textcolor{black}{\ensuremath{45.2\,\pm\,0.7}} \\\bottomrule
\end{tabular}

\end{table}

\paragraph{{Results under matched experience.}}
{PAVE achieves $98.9\%$ on LIBERO, $87.0\%$ on LIBERO-Plus, and $86.8\%/45.2\%$ on RoboTwin Clean/Random, improving on the closest matched control by $1.4$, $5.0$, and $4.3/4.5$ percentage points (Tables~\ref{tab:libero}--\ref{tab:robotwin}). Shared rollout data, initialization, actor updates, local prediction, and quality conditioning isolate trajectory-relative supervision. Fixed-pool factorial and routing controls in Appendix~\ref{app:interaction} further distinguish its effect from conditioning, target count, and mixed-quality filtering.}

\subsection{Quantitative Physical Evaluation on S1}
\label{sec:real}
We evaluate tabletop organization and beaker pouring on the \textbf{S1 robot platform}. Each task uses three independent training seeds, 120 demonstrations, and three physical evolution rounds of 50 rollouts each. The local-prediction + quality-conditioning control shares the demonstrations, cumulative rollouts, initialization protocol, and actor-update budget, omitting trajectory-relative targets. Table~\ref{tab:real} reports every seed and the mean $\pm$ sample SD across seeds, with 100 evaluation trials per seed and task. PAVE improves mean success by $5.7/7.3$ percentage points on organization/pouring. Appendix~\ref{app:realprotocol} details the protocol and statistical units.

\begin{table}[htbp]
\centering\scriptsize
\caption{Final S1 success (\%) after three evolution rounds. Local+QC retains local prediction and quality conditioning. Each seed has 100 trials per task and method. Mean $\pm$ sample SD summarizes the three training seeds (300 trials per task and method).}
\label{tab:real}
\begin{tabular}{@{}llcccc@{}}\toprule
Task & Method & Seed 1 & Seed 2 & Seed 3 & Mean $\pm$ SD \\\midrule
Tabletop organization & Local+QC & 80 & 79 & 81 & $80.0\,\pm\,1.0$ \\
 & PAVE & 86 & 89 & 82 & $85.7\,\pm\,3.5$ \\
Beaker pouring & Local+QC & 85 & 86 & 83 & $84.7\,\pm\,1.5$ \\
 & PAVE & 94 & 90 & 92 & $92.0\,\pm\,2.0$ \\\bottomrule
\end{tabular}

\end{table}

\begin{figure}[htbp]
\centering
\safeimage[width=1.0\textwidth]{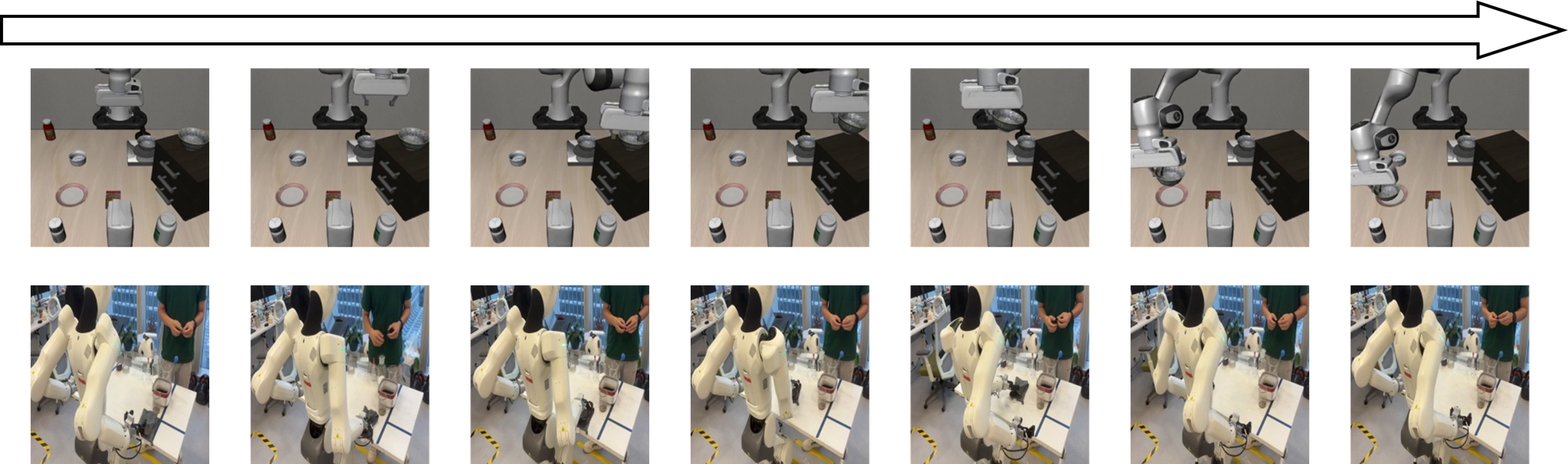}
\caption{Selected execution frames in simulation (top) and on the S1 platform (bottom). These qualitative examples illustrate execution over time; they do not determine the quantitative trial counts or success rates in Table~\ref{tab:real}.}
\label{fig:trajectory_vis}
\end{figure}

This comparison tests relative predictive supervision within physical evolution. As shown in Figure~\ref{fig:trajectory_vis}, representative simulation and S1 executions illustrate the evaluated manipulation behaviors. The evaluation concerns bounded autonomous task completion, not chemical reasoning, precise volumetric accuracy, or safety certification.

\subsection{Online Efficiency and Offline Cost}
The online actor requires no critic or transition-head evaluations. Profiling fixes token padding, camera count, action horizon, caching, dtype, and NVIDIA H200 hardware across methods (Table~\ref{tab:efficiency}); offline critic and predictive-branch costs are reported separately.
\begin{table}[htbp]
\centering\scriptsize
\caption{Matched online-efficiency profiling with two LIBERO views, $H=20$, four flow steps, batch size one, synchronized CUDA, and the same NVIDIA H200 GPU model.}
\label{tab:efficiency}
\begin{tabular}{@{}lccccc@{}}\toprule
Method & Online params (B) & p50 (ms) & p95 (ms) & GPU memory (GB) & Online critics \\\midrule
Matched JEPA-WAM & \textcolor{black}{\ensuremath{1.29}} & \textcolor{black}{\ensuremath{63.40}} & \textcolor{black}{\ensuremath{70.80}} & \textcolor{black}{\ensuremath{10.60}} & 0 \\
PAVE & \textcolor{black}{\ensuremath{1.29}} & \textcolor{black}{\ensuremath{63.60}} & \textcolor{black}{\ensuremath{71.00}} & \textcolor{black}{\ensuremath{10.60}} & 0 \\\bottomrule
\end{tabular}

\end{table}

Both actors use $1.29$B online parameters and $10.60$GB memory, with p50 latency $63.40/63.60$ms. In the profiled simulation setup, each round costs $31.20$ actor GPU-hours and $8.40$ across two critics, excluding collection/preparation (Appendix Table~\ref{tab:warmstart}); Appendix Table~\ref{tab:crossfit} controls total critic updates. Gains are at matched experience and actor updates, not equal offline compute.

\subsection{Temporal Alignment and Predictive Critic Ablations}
\label{sec:corecontrols}
\begingroup
\begin{table}[htbp]
\centering\small
\caption{Testing supervision roles, not additional experience. A fixes the data pool, initialization, and actor updates while varying relative targets and quality conditioning (QC). B fixes the pool and predictive-pair budget while varying reuse rules. Values are mean $\pm$ sample SD across five seeds.}
\label{tab:singlemulti}
\begin{tabular*}{\textwidth}{@{\extracolsep{\fill}}lcc@{}}\toprule
Training configuration & LIBERO-Plus & RoboTwin Random \\\midrule
\multicolumn{3}{@{}l}{\textit{A. Same-pool factorial: temporal targets $\times$ action conditioning}} \\
Local targets; unconditioned actions & $80.0\,\pm\,0.8$ & $38.4\,\pm\,1.1$ \\
Local + relative targets; unconditioned actions & $81.6\,\pm\,0.6$ & $39.8\,\pm\,0.9$ \\
Local targets; quality-conditioned actions & $82.0\,\pm\,1.0$ & $40.7\,\pm\,1.0$ \\
Local + relative targets; quality-conditioned actions & $87.0\,\pm\,0.7$ & $45.2\,\pm\,0.7$ \\\midrule
\multicolumn{3}{@{}l}{\textit{B. Fixed-pool experience-use controls}} \\
Success-only prediction; actions unchanged & $83.5\,\pm\,0.8$ & $41.8\,\pm\,1.3$ \\
Positive-rollout filtering for both branches & $82.8\,\pm\,0.9$ & $40.7\,\pm\,0.9$ \\
Stop predictive gradients during evolution & $83.1\,\pm\,0.6$ & $41.5\,\pm\,1.1$ \\
All-valid prediction; positive-only actions & $83.4\,\pm\,1.0$ & $41.9\,\pm\,0.7$ \\
PAVE: all-valid prediction; conditioned actions & $87.0\,\pm\,0.7$ & $45.2\,\pm\,0.7$ \\\bottomrule
\end{tabular*}
\end{table}
\endgroup
Table~\ref{tab:singlemulti}A isolates trajectory-relative alignment with quality conditioning. Relative targets gain $3.4/3.0$ points over fixed offsets at equal pair count (Appendix Table~\ref{tab:anchorplacement}); Table~\ref{tab:singlemulti}B tests continued prediction and mixed-quality reuse. With critic architecture, data, folds, and update budgets fixed, setting $\lambda_{\rm J}=0$ reduces success from $87.0/45.2$ to $85.5/44.3$ (Appendix Table~\ref{tab:valuefunction}). The $1.5/0.9$-point gain supports learning future latents through the critic's shared state representation. These ablations evaluate the actor's adaptive temporal alignment and the critic's action-consequence learning separately.

\PAVECaptionsNoColon
\section{Limitations and Discussion}
The actor's predictive representation is not an executable subgoal plan or an action-conditioned simulator; causal partial observation can blur unresolved futures. Its critic ranks recorded chunks under finite coverage and behavior-mixture bias. Binary conditions discard within-class quality and do not certify safe or improving actions.

S1 tests physical evolution with a matched temporal-supervision control; broader component isolation remains in simulation. More tasks and repeated training runs are needed. Extra offline training cost remains (Appendix Table~\ref{tab:warmstart}). Future work should study critic calibration and threshold stability, action-conditioned values with support checks, conditions for monotonic improvement, and longer real-robot evolution with explicit safety monitoring.

\section{Conclusion}
PAVE combines trajectory-adaptive predictive alignment, action-conditioned value learning, and quality-conditioned policy evolution. Multi-scale future targets train the actor across changing temporal spans, while the critic learns action consequences jointly with returns to guide offline chunk annotation. Matched actor and critic ablations isolate these mechanisms, and three-seed S1 evaluation shows gains under physical policy evolution. The resulting policy retains a direct observation-to-action path with all predictive heads and critics offline.

\clearpage
\paragraph{Ethics statement.}\label{app:ethics}
The physical study is restricted to bounded workspaces with emergency stopping, operator supervision, and task-specific speed/contact constraints. For the beaker-pouring task, reagent identity, hazard assessment, containment, and spill handling are treated as part of the documented evaluation protocol. Unknown-outcome training censoring and evaluation failure accounting are distinguished, and a safety intervention is counted according to the declared autonomous-task failure rule. The reported task-success rates should not be interpreted as safety certifications.

\paragraph{Reproducibility statement.}
Appendix~\ref{app:protocol} documents data provenance, train/test separation, cross-fitting, and statistics; Appendix~\ref{app:configuration} (Appendix Table~\ref{tab:hyperparams}) provides defaults. Reporting includes five-seed simulation means and sample SDs, individual S1 seed results with three-seed means and sample SDs, and NVIDIA H200 profiling.

\PAVECaptionsWithColon
\subsection*{AI use statement}
We used generative AI tools for language polishing, literature search and review, restructuring the presentation of the supervision formulation, and drafting manuscript revisions. OpenAI Codex also assisted with methodological explanations and feedback, including discussion of partial-observation limits, conditioning trade-offs, and proposed diagnostic protocols, as well as LaTeX preparation, bibliography consolidation, figure-label checks, and compilation. Experimental measurements were supplied by the authors; proposed diagnostics are identified as unreported. Compilation, cross-references, source consistency, and rendered layouts were checked during document preparation. The authors remain responsible for all scientific claims, references, and AI-assisted text or artifacts.

\bibliography{PAVE_all_references}

\begin{thebibliography}{34}
\providecommand{\natexlab}[1]{#1}
\providecommand{\url}[1]{\texttt{#1}}
\expandafter\ifx\csname urlstyle\endcsname\relax
  \providecommand{\doi}[1]{doi: #1}\else
  \providecommand{\doi}{doi: \begingroup \urlstyle{rm}\Url}\fi

\bibitem[Bellemare et~al.(2017)Bellemare, Dabney, and
  Munos]{bellemare2017distributional}
Marc~G. Bellemare, Will Dabney, and Remi Munos.
\newblock {A Distributional Perspective on Reinforcement Learning}.
\newblock In \emph{International Conference on Machine Learning}, pp.\
  449--458, 2017.

\bibitem[Black et~al.(2024)Black, Brown, Driess, Esmail, Equi, Finn, Fusai,
  Groom, Hausman, Ichter, et~al.]{black2024pi0}
Kevin Black, Noah Brown, Danny Driess, Adnan Esmail, Michael Equi, Chelsea
  Finn, Niccolo Fusai, Lachy Groom, Karol Hausman, Brian Ichter, et~al.
\newblock {{$\pi_0$}: A Vision-Language-Action Flow Model for General Robot
  Control}.
\newblock \emph{arXiv preprint arXiv:2410.24164}, 2024.
\newblock \doi{10.48550/arXiv.2410.24164}.

\bibitem[Chen et~al.(2025)Chen, Chen, Chen, Cai, Liu, Li, Liang, Lin, Ge, Gu,
  et~al.]{chen2025robotwin}
Tianxing Chen, Zanxin Chen, Baijun Chen, Zijian Cai, Yibin Liu, Zixuan Li,
  Qiwei Liang, Xianliang Lin, Yiheng Ge, Zhenyu Gu, et~al.
\newblock {RoboTwin 2.0: A Scalable Data Generator and Benchmark with Strong
  Domain Randomization for Robust Bimanual Robotic Manipulation}.
\newblock \emph{arXiv preprint arXiv:2506.18088}, 2025.
\newblock \doi{10.48550/arXiv.2506.18088}.

\bibitem[Chernozhukov et~al.(2016)Chernozhukov, Chetverikov, Demirer, Duflo,
  Hansen, Newey, and Robins]{chernozhukov2016crossfit}
Victor Chernozhukov, Denis Chetverikov, Mert Demirer, Esther Duflo, Christian
  Hansen, Whitney Newey, and James Robins.
\newblock {Double/Debiased Machine Learning for Treatment and Causal
  Parameters}.
\newblock \emph{arXiv preprint arXiv:1608.00060}, 2016.
\newblock \doi{10.48550/arXiv.1608.00060}.

\bibitem[Chi et~al.(2023)Chi, Xu, Feng, Cousineau, Du, Burchfiel, Tedrake, and
  Song]{chi2023diffusionpolicy}
Cheng Chi, Zhenjia Xu, Siyuan Feng, Eric Cousineau, Yilun Du, Benjamin
  Burchfiel, Russ Tedrake, and Shuran Song.
\newblock {Diffusion Policy: Visuomotor Policy Learning via Action Diffusion}.
\newblock In \emph{Robotics: Science and Systems}, 2023.

\bibitem[Fei et~al.(2025)Fei, Wang, Shi, Dai, Cai, Qian, Ji, He, Zhang, Fei,
  Fu, Gong, and Qiu]{fei2025liberoplus}
Senyu Fei, Siyin Wang, Junhao Shi, Zihao Dai, Jikun Cai, Pengfang Qian, Li~Ji,
  Xinzhe He, Shiduo Zhang, Zhaoye Fei, Jinlan Fu, Jingjing Gong, and Xipeng
  Qiu.
\newblock {LIBERO-Plus: In-depth Robustness Analysis of Vision-Language-Action
  Models}.
\newblock \emph{arXiv preprint arXiv:2510.13626}, 2025.
\newblock \doi{10.48550/arXiv.2510.13626}.

\bibitem[Fu et~al.(2026)Fu, Tan, Sun, Liu, Zheng, Xu, Zhu, Shen, and
  Xue]{fu2026lingbotvision}
Zelin Fu, Bin Tan, Changjiang Sun, Shaohui Liu, Kecheng Zheng, Yinghao Xu, Xing
  Zhu, Yujun Shen, and Nan Xue.
\newblock {Vision Pretraining for Dense Spatial Perception}.
\newblock \emph{arXiv preprint arXiv:2607.05247}, 2026.
\newblock \doi{10.48550/arXiv.2607.05247}.

\bibitem[Ghosh et~al.(2024)Ghosh, Walke, Pertsch, Black, Mees, Dasari, Hejna,
  Kreiman, Xu, Luo, Tan, Chen, Vuong, Xiao, Sanketi, Sadigh, Finn, and
  Levine]{ghosh2024octo}
Dibya Ghosh, Homer~Rich Walke, Karl Pertsch, Kevin Black, Oier Mees, Sudeep
  Dasari, Joey Hejna, Tobias Kreiman, Charles Xu, Jianlan Luo, You~Liang Tan,
  Lawrence~Yunliang Chen, Quan Vuong, Ted Xiao, Pannag~R Sanketi, Dorsa Sadigh,
  Chelsea Finn, and Sergey Levine.
\newblock {Octo: An Open-Source Generalist Robot Policy}.
\newblock In \emph{Proceedings of Robotics: Science and Systems}, Delft,
  Netherlands, 2024.
\newblock \doi{10.15607/RSS.2024.XX.090}.
\newblock URL \url{https://www.roboticsproceedings.org/rss20/p090.html}.

\bibitem[Hu et~al.(2022)Hu, Shen, Wallis, Allen-Zhu, Li, Wang, Wang, and
  Chen]{hu2022lora}
Edward~J. Hu, Yelong Shen, Phillip Wallis, Zeyuan Allen-Zhu, Yuanzhi Li, Shean
  Wang, Lu~Wang, and Weizhu Chen.
\newblock {LoRA: Low-Rank Adaptation of Large Language Models}.
\newblock In \emph{International Conference on Learning Representations}, 2022.

\bibitem[Kim et~al.(2024)Kim, Pertsch, Karamcheti, Xiao, Balakrishna, Nair,
  Rafailov, Foster, Lam, Sanketi, et~al.]{kim2024openvla}
Moo~Jin Kim, Karl Pertsch, Siddharth Karamcheti, Ted Xiao, Ashwin Balakrishna,
  Suraj Nair, Rafael Rafailov, Ethan Foster, Grace Lam, Pannag Sanketi, et~al.
\newblock {OpenVLA: An Open-Source Vision-Language-Action Model}.
\newblock In \emph{Conference on Robot Learning}, 2024.
\newblock arXiv:2406.09246.

\bibitem[Li et~al.(2026)Li, Zhang, Jin, Zeng, Zhuang, Tang, Lyu, and
  Wang]{li2026wav}
Runze Li, Hongyin Zhang, Junxi Jin, Qixin Zeng, Zifeng Zhuang, Yiqi Tang,
  Shangke Lyu, and Donglin Wang.
\newblock {World-Value-Action Model: Implicit Planning for
  Vision-Language-Action Systems}.
\newblock \emph{arXiv preprint arXiv:2604.14732}, 2026.
\newblock \doi{10.48550/arXiv.2604.14732}.

\bibitem[Lin et~al.(2026)Lin, He, Bao, Zhao, Li, Wang, Wang, Chi, and
  Zhang]{lin2026jepawam}
Yihan Lin, Jiawei He, Shifeng Bao, Chen Zhao, Yang Li, Xiaobo Wang, Yan Wang,
  Cheng Chi, and Jing Zhang.
\newblock {JEPA-WAM: Learning Vision-Language-Action Policies with
  Joint-Embedding World Modeling}.
\newblock \emph{arXiv preprint arXiv:2608.09381}, 2026.
\newblock \doi{10.48550/arXiv.2608.09381}.

\bibitem[Liu et~al.(2023)Liu, Zhu, Gao, Feng, Liu, Zhu, and
  Stone]{liu2023libero}
Bo~Liu, Yifeng Zhu, Chongkai Gao, Yihao Feng, Qiang Liu, Yuke Zhu, and Peter
  Stone.
\newblock {LIBERO: Benchmarking Knowledge Transfer for Lifelong Robot
  Learning}.
\newblock \emph{arXiv preprint arXiv:2306.03310}, 2023.
\newblock \doi{10.48550/arXiv.2306.03310}.

\bibitem[Liu et~al.(2026)Liu, Yang, Wang, Jiang, Chi, Xu, Ding, Chen, and
  Wang]{liu2026stagewam}
Xiao Liu, Yuguang Yang, Xi~Wang, Kai Jiang, Cheng Chi, Yong Xu, Wenchao Ding,
  Yilun Chen, and Yan Wang.
\newblock {StageWAM: Joint-Embedding Stage Prediction for World-Action Models
  in Robot Manipulation}.
\newblock \emph{arXiv preprint arXiv:2608.10780}, 2026.
\newblock \doi{10.48550/arXiv.2608.10780}.

\bibitem[Luo et~al.(2026)Luo, Zhang, Feng, Zheng, Xu, Xu, Xi, Fu, and
  Lu]{luo2026beingh07}
Hao Luo, Wanpeng Zhang, Yicheng Feng, Sipeng Zheng, Haiweng Xu, Chaoyi Xu,
  Ziheng Xi, Yuhui Fu, and Zongqing Lu.
\newblock {Being-H0.7: A Latent World-Action Model from Egocentric Videos}.
\newblock \emph{arXiv preprint arXiv:2605.00078}, 2026.
\newblock \doi{10.48550/arXiv.2605.00078}.

\bibitem[Ma et~al.(2026{\natexlab{a}})Ma, Zhou, Peng, Li, Gu, and
  Yan]{ma2026stpi}
Chuanhao Ma, Hanyu Zhou, Shihan Peng, Yan Li, Tao Gu, and Luxin Yan.
\newblock {{ST-$\pi$}: Structured SpatioTemporal VLA for Robotic Manipulation}.
\newblock \emph{arXiv preprint arXiv:2604.17880}, 2026{\natexlab{a}}.
\newblock \doi{10.48550/arXiv.2604.17880}.

\bibitem[Ma et~al.(2026{\natexlab{b}})Ma, Cai, Xu, Li, Yang, Tian, Cao, Zhu,
  Qiu, {Zhaxizhuoma}, Yang, Peng, Wei, Zhu, Jiang, Gao, Wang, Yuan, Li, Zhu,
  Wang, Ding, Pang, Zeng, Zhang, Zhou, Mu, Shen, and Zhang]{ma2026internvla15}
Haoxiang Ma, Junhao Cai, Xiaoxu Xu, Hao Li, Yuyin Yang, Yang Tian, Jiafei Cao,
  Hongrui Zhu, Zherui Qiu, {Zhaxizhuoma}, Yuqiang Yang, Jiaqi Peng, Xueyuan
  Wei, Yangkun Zhu, Jiahao Jiang, Xing Gao, Hanqing Wang, Feng Yuan, Kailin Li,
  Xueyue Zhu, Tai Wang, Yan Ding, Jiangmiao Pang, Jia Zeng, Jingjing Zhang,
  Bowen Zhou, Yao Mu, Chunhua Shen, and Weinan Zhang.
\newblock {InternVLA-A1.5: Unifying Understanding, Latent Foresight, and Action
  for Compositional Generalization}.
\newblock \emph{arXiv preprint arXiv:2607.04988}, 2026{\natexlab{b}}.
\newblock \doi{10.48550/arXiv.2607.04988}.

\bibitem[Mao et~al.(2026)Mao, Yu, Mao, Li, Hu, Lan, Zhu, and Chen]{mao2026arm}
Yiming Mao, Zixi Yu, Weixin Mao, Yinhao Li, Qirui Hu, Zihan Lan, Minzhao Zhu,
  and Hua Chen.
\newblock {{ARM}: Advantage Reward Modeling for Long-Horizon Manipulation}.
\newblock In \emph{Proceedings of the IEEE/CVF Conference on Computer Vision
  and Pattern Recognition Workshops}, pp.\  4468--4477, 2026.

\bibitem[Miao et~al.(2026)Miao, Feng, Wu, Lin, He, Li, and
  Long]{miao2026jepavla}
Shangchen Miao, Ningya Feng, Jialong Wu, Ye~Lin, Xu~He, Dong Li, and Mingsheng
  Long.
\newblock {JEPA-VLA: Video Predictive Embedding is Needed for VLA Models}.
\newblock \emph{arXiv preprint arXiv:2602.11832}, 2026.
\newblock \doi{10.48550/arXiv.2602.11832}.

\bibitem[Mur-Labadia et~al.(2026)Mur-Labadia, Muckley, Bar, Assran, Sinha,
  Rabbat, LeCun, Ballas, and Bardes]{murlabadia2026vjepa21}
Lorenzo Mur-Labadia, Matthew Muckley, Amir Bar, Mido Assran, Koustuv Sinha,
  Mike Rabbat, Yann LeCun, Nicolas Ballas, and Adrien Bardes.
\newblock {V-JEPA 2.1: Unlocking Dense Features in Video Self-Supervised
  Learning}.
\newblock \emph{arXiv preprint arXiv:2603.14482}, 2026.
\newblock \doi{10.48550/arXiv.2603.14482}.

\bibitem[Oquab et~al.(2023)Oquab, Darcet, Moutakanni, Vo, Szafraniec, Khalidov,
  Fernandez, Haziza, Massa, El-Nouby, et~al.]{oquab2023dinov2}
Maxime Oquab, Timoth\'ee Darcet, Th\'eo Moutakanni, Huy Vo, Marc Szafraniec,
  Vasil Khalidov, Pierre Fernandez, Daniel Haziza, Francisco Massa, Alaaeldin
  El-Nouby, et~al.
\newblock {DINOv2: Learning Robust Visual Features without Supervision}.
\newblock \emph{arXiv preprint arXiv:2304.07193}, 2023.
\newblock \doi{10.48550/arXiv.2304.07193}.

\bibitem[{Physical Intelligence} et~al.(2025{\natexlab{a}}){Physical
  Intelligence}, Amin, Aniceto, Balakrishna, Black, Conley, Connors, Darpinian,
  Dhabalia, et~al.]{physicalintelligence2025pistar06}
{Physical Intelligence}, Ali Amin, Raichelle Aniceto, Ashwin Balakrishna, Kevin
  Black, Ken Conley, Grace Connors, James Darpinian, Karan Dhabalia, et~al.
\newblock {{$\pi^{*}_{0.6}$}: a VLA That Learns From Experience}.
\newblock \emph{arXiv preprint arXiv:2511.14759}, 2025{\natexlab{a}}.
\newblock \doi{10.48550/arXiv.2511.14759}.

\bibitem[{Physical Intelligence} et~al.(2025{\natexlab{b}}){Physical
  Intelligence}, Black, Brown, Darpinian, Dhabalia, Driess, Esmail, Equi, Finn,
  et~al.]{physicalintelligence2025pi05}
{Physical Intelligence}, Kevin Black, Noah Brown, James Darpinian, Karan
  Dhabalia, Danny Driess, Adnan Esmail, Michael Equi, Chelsea Finn, et~al.
\newblock {{$\pi_{0.5}$}: a Vision-Language-Action Model with Open-World
  Generalization}.
\newblock \emph{arXiv preprint arXiv:2504.16054}, 2025{\natexlab{b}}.
\newblock \doi{10.48550/arXiv.2504.16054}.

\bibitem[Soleymanzadeh et~al.(2026)Soleymanzadeh, Zhang, Zhang, Zhang, Liang,
  She, and Zheng]{soleymanzadeh2026phrvla}
Davood Soleymanzadeh, Kaidi Zhang, Zhiyuan Zhang, Bihao Zhang, Xiao Liang,
  Yu~She, and Minghui Zheng.
\newblock {PHR-VLA: Planning Horizon Reasoning for Vision-Language-Action
  Models}.
\newblock \emph{arXiv preprint arXiv:2608.27609}, 2026.
\newblock \doi{10.48550/arXiv.2608.27609}.

\bibitem[Sun et~al.(2026)Sun, Zhang, Qi, Ren, Liu, Zhu, Sun, Jin, and
  Chen]{sun2026vlajepa}
Jingwen Sun, Wenyao Zhang, Zekun Qi, Shaojie Ren, Zezhi Liu, Hanxin Zhu,
  Guangzhong Sun, Xin Jin, and Zhibo Chen.
\newblock {VLA-JEPA: Enhancing Vision-Language-Action Model with Latent World
  Model}.
\newblock \emph{arXiv preprint arXiv:2602.10098}, 2026.
\newblock \doi{10.48550/arXiv.2602.10098}.

\bibitem[Wang et~al.(2026)Wang, Fan, Qiu, Shangguan, Fu, Fu, Seita, and
  Xue]{wang2026oflow}
Kuanning Wang, Ke~Fan, Chenhao Qiu, Zeyu Shangguan, Yuqian Fu, Yanwei Fu,
  Daniel Seita, and Xiangyang Xue.
\newblock {OFlow: Injecting Object-Aware Temporal Flow Matching for Robust
  Robotic Manipulation}.
\newblock \emph{arXiv preprint arXiv:2604.17876}, 2026.
\newblock \doi{10.48550/arXiv.2604.17876}.

\bibitem[Wang et~al.(2025)Wang, Gao, Fu, Li, Fortier, Mullins, Kolobov, and
  Guo]{wang2025lola}
Xiaofan Wang, Xingyu Gao, Jianlong Fu, Zuolei Li, Dean Fortier, Galen Mullins,
  Andrey Kolobov, and Baining Guo.
\newblock {{LoLA}: Long Horizon Latent Action Learning for General Robot
  Manipulation}.
\newblock \emph{arXiv preprint arXiv:2512.20166}, 2025.
\newblock \doi{10.48550/arXiv.2512.20166}.

\bibitem[Wu et~al.(2026)Wu, Zhang, Tan, Wang, and Zhang]{wu2026flowpro}
Yihao Wu, He~Zhang, Junbo Tan, Xueqian Wang, and Zhengyou Zhang.
\newblock {FlowPRO: Reward-Free Reinforced Fine-Tuning of Flow-Matching VLAs
  via Proximalized Preference Optimization}.
\newblock \emph{arXiv preprint arXiv:2606.05468}, 2026.
\newblock \doi{10.48550/arXiv.2606.05468}.

\bibitem[Yan et~al.(2026)Yan, Li, Zhu, Wang, Shou, Miao, Pang, Hong, and
  Guo]{yan2026redflow}
Zhengyang Yan, Junhao Li, Fangqi Zhu, Zijun Wang, Quanxin Shou, Yikun Miao,
  Xiaoyi Pang, Zicong Hong, and Song Guo.
\newblock {RedFlow: Redirect Failure into Action-Level Corrections for
  Flow-matching VLA Policy}.
\newblock \emph{arXiv preprint arXiv:2607.27782}, 2026.
\newblock \doi{10.48550/arXiv.2607.27782}.

\bibitem[Yang et~al.(2024)Yang, Yang, Zhang, Hui, Zheng, Yu, Li, Liu, Huang,
  Wei, et~al.]{yang2024qwen25}
An~Yang, Baosong Yang, Beichen Zhang, Binyuan Hui, Bo~Zheng, Bowen Yu,
  Chengyuan Li, Dayiheng Liu, Fei Huang, Haoran Wei, et~al.
\newblock {Qwen2.5 Technical Report}.
\newblock \emph{arXiv preprint arXiv:2412.15115}, 2024.
\newblock \doi{10.48550/arXiv.2412.15115}.

\bibitem[Yang et~al.(2026)Yang, Wang, Wu, Liu, Yan, Du, Yue, Zhang, Wang, Liu,
  Qi, Chen, Shan, and Yao]{yang2026aloe}
Rushuai Yang, Hecheng Wang, Zhichao Wu, Chiming Liu, Xiaohan Yan, Xuan Du,
  Shuoyu Yue, Chuheng Zhang, Yunlong Wang, Yongcheng Liu, Lizhe Qi, Yi~Chen,
  Wei Shan, and Maoqing Yao.
\newblock {{ALOE}: Action-Level Off-Policy Evaluation for
  Vision-Language-Action Model Post-Training}.
\newblock \emph{arXiv preprint arXiv:2602.12691}, 2026.
\newblock \doi{10.48550/arXiv.2602.12691}.

\bibitem[Yuan et~al.(2026)Yuan, Dong, Liu, and Zhao]{yuan2026fastwam}
Tianyuan Yuan, Zibin Dong, Yicheng Liu, and Hang Zhao.
\newblock {Fast-WAM: Do World Action Models Need Test-time Future Imagination?}
\newblock \emph{arXiv preprint arXiv:2603.16666}, 2026.
\newblock \doi{10.48550/arXiv.2603.16666}.

\bibitem[Zhai et~al.(2025)Zhai, Zhao, Yu, and Hao]{zhai2025vfp}
Xuanran Zhai, Qianyou Zhao, Qiaojun Yu, and Ce~Hao.
\newblock {VFP: Variational Flow-Matching Policy for Multi-Modal Robot
  Manipulation}.
\newblock \emph{arXiv preprint arXiv:2508.01622}, 2025.
\newblock \doi{10.48550/arXiv.2508.01622}.

\bibitem[Zitkovich et~al.(2023)Zitkovich, Yu, Xu, Xu, Xiao, Xia, Wu, Wohlhart,
  Welker, Wahid, Vuong, Vanhoucke, Tran, Soricut, Singh, Singh, Sermanet,
  Sanketi, Salazar, Ryoo, Reymann, Rao, Pertsch, Mordatch, Michalewski, Lu,
  Levine, Lee, Lee, Leal, Kuang, Kalashnikov, Julian, Joshi, Irpan, Ichter,
  Hsu, Herzog, Hausman, Gopalakrishnan, Fu, Florence, Finn, Dubey, Driess,
  Ding, Choromanski, Chen, Chebotar, Carbajal, Brown, Brohan, Arenas, and
  Han]{zitkovich2023rt2}
Brianna Zitkovich, Tianhe Yu, Sichun Xu, Peng Xu, Ted Xiao, Fei Xia, Jialin Wu,
  Paul Wohlhart, Stefan Welker, Ayzaan Wahid, Quan Vuong, Vincent Vanhoucke,
  Huong Tran, Radu Soricut, Anikait Singh, Jaspiar Singh, Pierre Sermanet,
  Pannag~R. Sanketi, Grecia Salazar, Michael~S. Ryoo, Krista Reymann, Kanishka
  Rao, Karl Pertsch, Igor Mordatch, Henryk Michalewski, Yao Lu, Sergey Levine,
  Lisa Lee, Tsang-Wei~Edward Lee, Isabel Leal, Yuheng Kuang, Dmitry
  Kalashnikov, Ryan Julian, Nikhil~J. Joshi, Alex Irpan, Brian Ichter, Jasmine
  Hsu, Alexander Herzog, Karol Hausman, Keerthana Gopalakrishnan, Chuyuan Fu,
  Pete Florence, Chelsea Finn, Kumar~Avinava Dubey, Danny Driess, Tianli Ding,
  Krzysztof~Marcin Choromanski, Xi~Chen, Yevgen Chebotar, Justice Carbajal,
  Noah Brown, Anthony Brohan, Montserrat~Gonzalez Arenas, and Kehang Han.
\newblock {RT-2: Vision-Language-Action Models Transfer Web Knowledge to
  Robotic Control}.
\newblock In \emph{Proceedings of The 7th Conference on Robot Learning}, volume
  229 of \emph{Proceedings of Machine Learning Research}, pp.\  2165--2183.
  PMLR, 2023.
\newblock URL \url{https://proceedings.mlr.press/v229/zitkovich23a.html}.

\end{thebibliography}
\bibliographystyle{iclr2027_conference}
\clearpage
\appendix
\section*{Appendix}

\section{Detailed Experimental Protocol}
\label{app:protocol}
\subsection{Default Model and Training Configuration}
\label{app:configuration}
Table~\ref{tab:hyperparams} summarizes the actor, dual-head critic, and policy-evolution configuration. Ablations override the settings explicitly identified in their descriptions; published reference results retain their source configurations. Simulation rollout budgets are per training group per seed; S1 budgets are per task. Critic-update counts are per fold-specific critic.

\begin{table}[ht]
\centering\scriptsize
\caption{Default actor, dual-head critic, and policy-evolution configuration.}
\label{tab:hyperparams}
\begin{tabularx}{\textwidth}{@{}lX@{}}\toprule
Component & Configuration \\\midrule
Actor vision & {Frozen V-JEPA~2.1 ViT-L/16; $384^2$ inputs; $576\times1024$ per view; $N_v=576n_v$; frozen projector to width 896} \\
Actor predictor & Qwen2.5-0.5B, width $896$; $64$ action placeholders \\
Adaptation / decoder & LoRA rank $32$, $\alpha=64$, dropout $0.1$; $16$-layer DiT \\
Learnable future queries & $32$ learnable query tokens; separate from the predictive targets \\
Action interface & LIBERO: two views, $d_a=7$; RoboTwin: \textcolor{black}{three views, $d_a=14$}; S1: \textcolor{black}{three RGB views, $d_a=14$} \\
Controlled horizon & $H=N=20$; four Euler steps; execution horizon $20$ \\
Predictive supervision & $\delta=31$; $\rho=\{.25,.50,.75,1\}$; both loss weights $0.5$ \\
Critic & V-JEPA~2.1 + Gemma; pooled raw-feature concatenation; $(D_v+D_l)\to1024\to512$ fusion; separate return/JEPA heads; $201$ bins on $[-1,0]$ \\
Gemma implementation & \textcolor{black}{Gemma-2-2B, frozen, width $2304$; checkpoint tokenizer; 256-bin normalized robot-state discretization} \\
Cross-fitting & Episode-disjoint; $F=\textcolor{black}{2}$; same excluded-fold critic for both values \\
Labels / replay & Within-task top $30\%$; label dropout $0.3$; demos:evolution $3{:}1$ \\
Evolution (simulation) & Three rounds; $96$ new episodes per training group per seed per round \\
S1 evolution & Three rounds; $50$ new physical episodes per task per round; $150$ per task total \\
Optimization & Base actor $60{,}000$ updates; actor $30{,}000$ updates/evolution round; $8{,}000$ per fold-specific critic per round \\
Global batches & Actor $128$; pooled-critic batch \textcolor{black}{$240$} \\
Statistics & Simulation: mean and sample SD across five independent training seeds; S1: individual results, mean and sample SD across three independent training seeds \\\bottomrule
\end{tabularx}
\end{table}

\subsection{Data Provenance, Counts, and Benchmark Compatibility}
The run-manifest schema records the task list, initial-state identifiers, training episodes, collection-policy lineage, encoder/policy checkpoints, code commit, preprocessing, action units, and evaluator version. Published numbers retain their source attribution; source citations alone do not establish a match to our training or evaluation configuration. Table~\ref{tab:datacounts} summarizes the data scale of the controlled runs.
\begin{table}[ht]
\centering\scriptsize
\caption{Demonstration-data inventory for adaptation. LIBERO-Plus has no separate training pool. Additional deployment episodes are counted separately in Appendix~\ref{app:realprotocol} and Table~\ref{tab:rounds}.}
\label{tab:datacounts}
\begin{tabularx}{\textwidth}{@{}lXXXX@{}}\toprule
Track & Demos / task & Recorded demo frames & Valid actor chunks & Valid local / MH pairs \\\midrule
LIBERO clean & \textcolor{black}{50} & \textcolor{black}{336,575} & \textcolor{black}{298,575} & \textcolor{black}{274,575 / 1,194,300} \\
LIBERO-Plus test & No OOD training & Not training frames & Not training chunks & Not training pairs \\
RoboTwin Clean-20 & \textcolor{black}{50} & \textcolor{black}{414,687} & \textcolor{black}{395,687} & \textcolor{black}{383,687 / 1,582,748} \\
S1 organization & \textcolor{black}{120} & \textcolor{black}{36,418} & \textcolor{black}{34,138} & \textcolor{black}{32,698 / 136,552} \\
S1 beaker pouring & \textcolor{black}{120} & \textcolor{black}{32,852} & \textcolor{black}{30,572} & \textcolor{black}{29,132 / 122,288} \\\bottomrule
\end{tabularx}
\end{table}
Table~\ref{tab:datacounts} distinguishes recorded frames, valid actor chunks, and valid local/multi-horizon target pairs. Validity masks exclude padded, censored, or provenance-inconsistent windows. Repeated optimizer sampling and augmentation do not create additional recorded episodes. The local and multi-horizon pair counts are separate supervision counts rather than additional deployment experience.

\begin{table}[ht]
\centering\scriptsize
\caption{Protocol comparison between the published JEPA-WAM configuration \citep{lin2026jepawam} and the controlled PAVE setting.}
\label{tab:protocolcompare}
\begin{tabularx}{\textwidth}{@{}l>{\raggedright\arraybackslash}X>{\raggedright\arraybackslash}X@{}}\toprule
Item & Published JEPA-WAM & PAVE controlled setting \\\midrule
LIBERO action chunk & $H=8$, velocity prediction & $H=20$, velocity prediction \\
RoboTwin action chunk & $H=50$, $14$-D, $x$-prediction & \textcolor{black}{$H=20$, $14$-D, velocity prediction} \\
RoboTwin views / grouping & Three views; two groups of ten tasks & \textcolor{black}{Three views; same two groups of ten tasks} \\
Policy training & LIBERO $60{,}000$ updates; batch $128$ & \textcolor{black}{Base $60{,}000$ + $3\times30{,}000$ evolution updates; batch $128$} \\
Evolution experience & Clean demonstrations in published track & \textcolor{black}{96 disjoint clean rollouts/round/group/seed; 288 cumulative} \\
Generalization evaluation & No LIBERO-Plus fine-tuning; Clean-trained RoboTwin & No test-domain trajectories in any training or selection stage \\\bottomrule
\end{tabularx}
\end{table}
Table~\ref{tab:protocolcompare} contrasts the published source protocol with the controlled PAVE interface. The local demonstration-only JEPA-WAM baseline shares that interface and the clean demonstration budget. {The experience-based comparisons, including the local-prediction + quality-conditioning control, unlabeled BC, and the factorial/routing controls, additionally share the cumulative demonstration-plus-rollout pool and actor-update budget.} Thus published reference margins, end-to-end evolution gains, and same-data training-objective gains answer different questions. These simulation rollout counts are per training group per seed: $96$ per round and $288$ after three rounds. S1 instead uses $50$ per task per round, totaling $150$ per task.

\subsection{Episode-Level Cross-Fitting and Calibration Boundaries}
The default uses episode-disjoint two-fold cross-fitting ($F=2$). Fold membership is assigned by original episode identifier and retained as rounds accumulate; frames, chunks, and derived segments from one episode remain together. Related resets or derived copies are grouped to prevent near-duplicate leakage. Stratification uses task, collection round, and available outcome classes where the episode counts permit it.

For fold $f$, fit both return and JEPA objectives on $\mathcal D_{\leq k}\setminus\mathcal F_f$. Frozen pretrained encoders may be reused; an encoder fine-tuned on the complete rollout pool may not subsequently be described as fold-excluded. Fit normalization, $\lambda_{\rm J}$, early stopping, temperature, and any learned score transformation using a nested episode split of the training complement. Produce every held-out chunk's starting and bootstrap values using the same critic $\phi_k^{(-f)}$. Store the fold, model hash, normalization hash, advantage, and final condition alongside the immutable trajectory. The actor may train on all out-of-fold-labeled episodes.

Only complete, finite, provenance-consistent actor chunks enter action training and quality ranking. A true terminal transition uses no bootstrap value; padding and replicated endpoint frames are never counted as real actions. For $N\neq H$, both the executed action chunk and the reward/bootstrap interval must satisfy their separate validity masks. An unknown-outcome interruption is not assigned a terminal failure return simply because recording stopped. Its observed within-segment transitions can still support prediction when their physical/provenance validity is known.

\subsection{Statistical Units and Model Selection}
The simulation experiments use five independently trained seeds per retained configuration. We first compute task/perturbation means within each seed and then report their mean and sample standard deviation across seeds. Aggregate standard deviations are computed from seed-level aggregate scores rather than by averaging column standard deviations. A seed, rather than an individual rollout or critic fold, is the independent training replicate. Matched policies reuse held-out initializations. A macro-average across LIBERO-Plus and RoboTwin Random is used only as a diagnostic summary, not as an official benchmark score. The S1 experiments use three independent training seeds per task and method, each evaluated on 100 trials. We report all seed-level rates and their mean and sample SD, using three training replicates as the statistical sample size.

Defaults and checkpoints are selected using training-side validation only. Sensitivity grids are fixed before inspecting official test outcomes; post hoc selection of a best test row is not used to redefine the default. For matched method comparisons, the common cumulative experience pool is fixed before fitting the compared actors. Table~\ref{tab:rounds} reports PAVE's own collection--update trajectory as data and training accumulate. It is not a comparison of methods independently collecting their own rollouts; the main method comparisons and S1 control support shared-pool attribution.

\subsection{S1 Validation Protocol and Failure Accounting}
\label{app:realprotocol}
\paragraph{Physical training and evolution.}
Tabletop-organization and beaker-pouring experiments each use three independent training seeds per method. Matched actors share the initialization protocol and first adapt with 120 demonstrations per task. Each task's shared training pool is built through three physical collection--annotation--update rounds: 50 new deployment episodes are added per round, giving cumulative task pools of 50, 100, and 150 episodes. Across both tasks this is 300 deployment episodes, separate from the 240 demonstrations. Offline critics fit cumulative mixed-quality deployment experience and assign out-of-fold chunk scores; valid observed transitions support predictive learning, and both quality conditions support action learning. The actor is updated with demonstrations and annotated rollouts. This is an executed physical evolution loop, not only demonstration adaptation or offline simulation. The frame/chunk/pair inventory in Table~\ref{tab:datacounts} describes the demonstration pool; it is not a cumulative ledger of these additional deployment rollouts.

\paragraph{Matched physical control.}
Across three independent training seeds, the local-prediction + quality-conditioning control and PAVE share each task's 120 demonstrations, cumulative $3\times50$ physical rollouts, initialization protocol, and actor-update budget. The control removes trajectory-relative targets while retaining local prediction and critic-derived conditioning. After round 3, organization success is $(86,89,82)\%$ for PAVE and $(80,79,81)\%$ for Local+QC, giving $85.7\pm3.5\%$ and $80.0\pm1.0\%$ across seeds. Pouring success is $(94,90,92)\%$ and $(85,86,83)\%$, giving $92.0\pm2.0\%$ and $84.7\pm1.5\%$. The differences in seed means are $5.7$ and $7.3$ percentage points, respectively (Table~\ref{tab:real}).

\paragraph{Evaluation units and separation.}
Each method is tested on 100 trials per seed and task, spanning five initial-configuration groups with twenty trials each. Three independently trained seeds yield 300 trials per task and method, 600 per method across both tasks, and 1,200 across the two methods. These trials are separate from training collection, critic fitting, and actor updates. Trial order across configurations is randomized; session identifiers, scene resets, checkpoints, and interventions are recorded.

\paragraph{Uncertainty across training seeds.}
For seed-level success rates $p_1,p_2,p_3$ in percent, we report $\bar p=\frac{1}{3}\sum_i p_i$ and $s_p=\sqrt{\frac{1}{2}\sum_i(p_i-\bar p)^2}$. The SD measures variability among independently trained and evaluated policies, incorporating finite-trial evaluation variation; it is not a standard error or confidence interval. Equal trial counts make the seed mean numerically equal to pooled success, but the 300 trials do not constitute 300 independent training replicates. Reporting all three rates exposes variation that a pooled binomial interval would omit. With three seeds, these are descriptive uncertainty estimates; we do not claim statistical significance from the aggregate rates.

For organization, object-to-region assignments are fixed before evaluation. For pouring, the target region, minimum transfer requirement, spill criterion, and final beaker state are fixed in advance. Time to completion, transferred amount where measurable, spill amount, interventions, and failures are logged. Task horizons and tolerances follow the application requirements. Safe, institutionally approved materials and task-specific hazard handling define the protocol.

A rescue or task-related safety stop counts as an autonomous-task failure under the evaluation protocol. Hardware-invalid trials follow a separate predefined invalidation rule and exclusion log. This differs from training: an interrupted episode with an unknown return is not given an invented terminal failure target. Corrupted observations and invalid action intervals are masked, and no training pair crosses an intervention boundary. Physically valid observed segments may still support prediction. These rules separate useful task failures from invalid recordings without treating the critic as an online safety mechanism.

\section{Additional Ablations and Empirical Analyses}
\label{app:ablations}
Unless stated otherwise, matched actor controls use identical views, chunk interface, training seeds, and evaluator manifests. ``RT Random'' always denotes the same $20$-task source-aligned evaluation, not an average of isolated perturbation factors. Repeated full-PAVE reference rows use the same means and sample standard deviations as Tables~\ref{tab:libero}--\ref{tab:robotwin}.

\subsection{Policy Evolution and Advantage Conditioning}
\label{app:evolution_ablations}
\begin{table}[ht]
\centering\scriptsize
\caption{Policy-evolution ablation with the JEPA-assisted critic. Episode counts are per training group per seed. The base actor is the local+multi-horizon demonstration-only actor from Table~\ref{tab:lossablation}.}
\label{tab:rounds}
\begin{tabular}{@{}lcccc@{}}\toprule
Policy & New episodes & Cumulative episodes & LIBERO-Plus & RT Random \\\midrule
Base actor & 0 & 0 & \textcolor{black}{\ensuremath{81.2\,\pm\,0.7}} & \textcolor{black}{\ensuremath{39.0\,\pm\,1.3}} \\
Round 1 & \textcolor{black}{96} & \textcolor{black}{96} & \textcolor{black}{\ensuremath{83.6\,\pm\,1.2}} & \textcolor{black}{\ensuremath{40.8\,\pm\,0.9}} \\
Round 2 & \textcolor{black}{96} & \textcolor{black}{192} & \textcolor{black}{\ensuremath{85.9\,\pm\,1.1}} & \textcolor{black}{\ensuremath{42.2\,\pm\,0.8}} \\
Round 3 & \textcolor{black}{96} & \textcolor{black}{288} & \textcolor{black}{\ensuremath{87.0\,\pm\,0.7}} & \textcolor{black}{\ensuremath{45.2\,\pm\,0.7}} \\\bottomrule
\end{tabular}
\end{table}
{Table~\ref{tab:rounds} evaluates the JEPA-assisted critic within policy evolution. From the base actor through rounds 1--3, LIBERO-Plus success increases from $81.2$ to $83.6$, $85.9$, and $87.0$; RT Random rises from $39.0$ to $40.8$, $42.2$, and $45.2$. Each round jointly trains the critic on cumulative experience, recomputes quality labels, and updates the actor. The local-prediction + quality-conditioning control provides the matched actor comparison in Tables~\ref{tab:libero}--\ref{tab:robotwin}; Tables~\ref{tab:advantage} and~\ref{tab:interaction} compare unlabeled behavior cloning and conditional extraction on a fixed cumulative pool.}

\begin{table}[ht]
\centering\scriptsize
\caption{Conditioning controls. ``No evolution'' is the same base actor as Table~\ref{tab:rounds}, evaluated under its positive condition. Other rows share the cumulative pool. The last three rows change only the test condition of one trained actor.}
\label{tab:advantage}
\begin{tabular}{@{}llccc@{}}\toprule
Training / test condition & Data & LIBERO-Plus & RT Random & Diagnostic mean \\\midrule
No evolution / positive & Demos & \textcolor{black}{\ensuremath{81.2\,\pm\,0.8}} & \textcolor{black}{\ensuremath{39.0\,\pm\,0.9}} & \textcolor{black}{\ensuremath{60.1\,\pm\,0.4}} \\
Unlabeled multi-round BC / null & Demos + rollouts & \textcolor{black}{\ensuremath{81.6\,\pm\,0.8}} & \textcolor{black}{\ensuremath{39.8\,\pm\,1.0}} & \textcolor{black}{\ensuremath{60.7\,\pm\,0.3}} \\
ACP / null & Same cumulative pool & \textcolor{black}{\ensuremath{82.9\,\pm\,0.8}} & \textcolor{black}{\ensuremath{41.6\,\pm\,1.1}} & \textcolor{black}{\ensuremath{62.3\,\pm\,0.8}} \\
ACP / negative & Same cumulative pool & \textcolor{black}{\ensuremath{66.8\,\pm\,1.4}} & \textcolor{black}{\ensuremath{24.7\,\pm\,1.5}} & \textcolor{black}{\ensuremath{45.8\,\pm\,0.8}} \\
ACP / positive & Same cumulative pool & \textcolor{black}{\ensuremath{87.0\,\pm\,0.7}} & \textcolor{black}{\ensuremath{45.2\,\pm\,0.7}} & \textcolor{black}{\ensuremath{66.1\,\pm\,0.5}} \\\bottomrule
\end{tabular}
\end{table}
Conditioning can separate behavioral modes without removing all negative examples from training. Null-condition dropout provides an unconditioned data-distribution reference; it is not an uncertainty label assigned by the critic. A positive-versus-negative test contrast alone would establish that the condition changes behavior, not that the labels are accurate. Shuffled-label and positive-only-training controls in Section~\ref{app:interaction} address that distinction.

\subsection{Visual Encoder Choice}
\label{app:encoders}
\begin{figure}[ht]
\centering
\safeimage[width=.88\textwidth,height=1.8in,keepaspectratio]{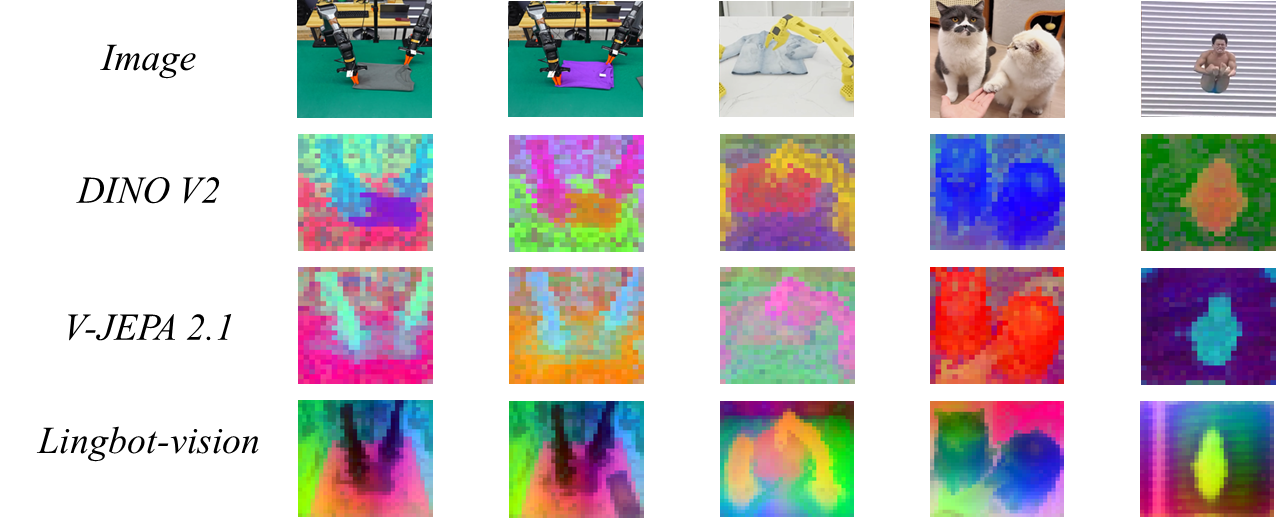}
\caption{Qualitative encoder response maps for the same displayed inputs. Rows show DINOv2, V-JEPA~2.1, and LingBot-Vision. The maps illustrate spatial response patterns; quantitative encoder comparisons are given in Table~\ref{tab:encoder}, rather than inferred from heatmap colors.}
\label{fig:encoder_vis}
\end{figure}

\begin{table}[ht]
\centering\scriptsize
\caption{Single-factor actor-perception ablation. The V-JEPA~2.1 target encoder and pooled V-JEPA~2.1+Gemma critic are fixed; only the current actor visual encoder and required projector are changed.}
\label{tab:encoder}
\begin{tabular}{@{}lcccc@{}}\toprule
Current actor encoder & LIBERO & LIBERO-Plus & RT Random & Training steps/s \\\midrule
DINOv2 & \textcolor{black}{\ensuremath{97.5\,\pm\,0.5}} & \textcolor{black}{\ensuremath{82.0\,\pm\,1.0}} & \textcolor{black}{\ensuremath{40.4\,\pm\,0.9}} & \textcolor{black}{\ensuremath{1.27}} \\
LingBot-Vision & \textcolor{black}{\ensuremath{97.8\,\pm\,0.5}} & \textcolor{black}{\ensuremath{83.1\,\pm\,0.6}} & \textcolor{black}{\ensuremath{41.7\,\pm\,1.0}} & \textcolor{black}{\ensuremath{1.10}} \\
V-JEPA~2.1 & \textcolor{black}{\ensuremath{98.9\,\pm\,0.2}} & \textcolor{black}{\ensuremath{87.0\,\pm\,0.7}} & \textcolor{black}{\ensuremath{45.2\,\pm\,0.7}} & \textcolor{black}{\ensuremath{1.00}} \\\bottomrule
\end{tabular}
\end{table}
Table~\ref{tab:encoder} changes the current-observation encoder and required projector while fixing the V-JEPA~2.1 target encoder and pooled critic. V-JEPA~2.1 gives the highest reported success means, whereas the other actor encoders have higher reported training-step rates. This is an actor-interface comparison, separate from changing the predictive teacher or critic. Figure~\ref{fig:encoder_vis} provides qualitative response maps; the success-rate comparison relies on the controlled evaluations.

\subsection{Number, Placement, and Budget of Future Anchors}
\label{app:horizons}
\begin{table}[ht]
\centering\scriptsize
\caption{Sensitivity to the number and placement of trajectory-relative predictive anchors. The local target and quality conditioning are retained in every row.}
\label{tab:numhorizons}
\begin{tabular}{@{}clcc@{}}\toprule
$K$ & Relative fractions & LIBERO-Plus & RT Random \\\midrule
\textcolor{black}{1} & \textcolor{black}{$\{1\}$} & \textcolor{black}{\ensuremath{82.1\,\pm\,0.8}} & \textcolor{black}{\ensuremath{40.9\,\pm\,0.9}} \\
\textcolor{black}{2} & \textcolor{black}{$\{.5,1\}$} & \textcolor{black}{\ensuremath{83.5\,\pm\,1.5}} & \textcolor{black}{\ensuremath{42.0\,\pm\,1.3}} \\
\textcolor{black}{3} & \textcolor{black}{$\{1/3,2/3,1\}$} & \textcolor{black}{\ensuremath{84.0\,\pm\,0.9}} & \textcolor{black}{\ensuremath{42.6\,\pm\,1.1}} \\
4 & $\{.25,.5,.75,1\}$ & \textcolor{black}{\ensuremath{87.0\,\pm\,0.7}} & \textcolor{black}{\ensuremath{45.2\,\pm\,0.7}} \\
\textcolor{black}{6} & \textcolor{black}{$\{j/6:j=1,\ldots,6\}$} & \textcolor{black}{\ensuremath{85.7\,\pm\,1.1}} & \textcolor{black}{\ensuremath{43.0\,\pm\,0.6}} \\
\textcolor{black}{8} & \textcolor{black}{$\{j/8:j=1,\ldots,8\}$} & \textcolor{black}{\ensuremath{84.4\,\pm\,0.8}} & \textcolor{black}{\ensuremath{42.5\,\pm\,1.3}} \\\bottomrule
\end{tabular}
\end{table}
\begin{table}[ht]
\centering\scriptsize
\caption{Target-count-matched schedule controls. Each row uses four auxiliary targets plus the local objective and an equal target-pair sampling budget.}
\label{tab:anchorplacement}
\begin{tabular}{@{}lcc@{}}\toprule
Schedule & LIBERO-Plus & RT Random \\\midrule
Fixed offsets \textcolor{black}{$\{16,32,64,128\}$}, endpoint-clipped & \textcolor{black}{\ensuremath{83.6\,\pm\,0.8}} & \textcolor{black}{\ensuremath{42.2\,\pm\,1.3}} \\
\textcolor{black}{Four sorted uniform random relative fractions} & \textcolor{black}{\ensuremath{84.2\,\pm\,0.8}} & \textcolor{black}{\ensuremath{42.8\,\pm\,1.3}} \\
Trajectory-relative fixed fractions & \textcolor{black}{\ensuremath{87.0\,\pm\,0.7}} & \textcolor{black}{\ensuremath{45.2\,\pm\,0.7}} \\\bottomrule
\end{tabular}
\end{table}
\paragraph{Why more anchors need not improve performance.}
Table~\ref{tab:numhorizons} gives $84.0/42.6$ for $K=3$, $87.0/45.2$ for $K=4$, $85.7/43.0$ for $K=6$, and $84.4/42.5$ for $K=8$ on LIBERO-Plus/RT Random. For equally spaced fractions $\rho_j=j/K$, Equation~\ref{eq:horizonindex} gives adjacent-anchor spacing $\Delta h_t\approx(T-t)/K$ before rounding. Thus the temporal interval changes along each recorded rollout: anchors span broad changes early and contract as the rollout approaches its endpoint. Increasing $K$ further shortens this interval at every position.

With fewer anchors, intermediate changes receive coarser coverage. Four anchors balance coverage and temporal resolution in these experiments. A plausible explanation for the decline at $K=6$ is that denser late-trajectory targets increasingly emphasize small local details and minor state differences. Through shared-representation training, this emphasis can encourage overly fine action corrections and slight jitter near completion, reducing success. Masking zero-distance pairs and downweighting duplicate indices prevents exact repetition from receiving extra weight, but does not remove highly similar targets at distinct nearby times. The success rates support this coverage--resolution trade-off; jitter is a mechanistic interpretation, not a separately quantified outcome. The anchors remain training-only temporal references, rather than semantic subgoals or online inputs.

At an equal target-pair sampling budget, Table~\ref{tab:anchorplacement} gives $87.0/45.2$ for fixed relative fractions, $83.6/42.2$ for fixed offsets, and $84.2/42.8$ for random relative fractions. This supports trajectory-relative coverage under the tested schedules without attributing the difference to extra target evaluations.

\subsection{Loss Components and Data Regimes}
\label{app:loss}
\begin{table}[ht]
\centering\scriptsize
\caption{Loss/evolution controls. Rows without quality conditioning use demonstrations only; rows with quality conditioning also use cumulative rollouts. Thus this table is not, by itself, a same-data interaction test. Table~\ref{tab:interaction} supplies that control.}
\label{tab:lossablation}
\begin{tabular}{@{}cccccc@{}}\toprule
$\mathcal L_{\rm FM}$ & $\mathcal L_{\rm local}$ & $\mathcal L_{\rm MH}$ & ACP & LIBERO-Plus & RT Random \\\midrule
$\checkmark$ & & & & \textcolor{black}{\ensuremath{76.9\,\pm\,0.8}} & \textcolor{black}{\ensuremath{32.0\,\pm\,1.5}} \\
$\checkmark$ & $\checkmark$ & & & \textcolor{black}{\ensuremath{79.0\,\pm\,1.0}} & \textcolor{black}{\ensuremath{36.7\,\pm\,1.7}} \\
$\checkmark$ & & $\checkmark$ & & \textcolor{black}{\ensuremath{80.0\,\pm\,0.8}} & \textcolor{black}{\ensuremath{37.8\,\pm\,1.0}} \\
$\checkmark$ & $\checkmark$ & $\checkmark$ & & \textcolor{black}{\ensuremath{81.2\,\pm\,1.1}} & \textcolor{black}{\ensuremath{39.0\,\pm\,0.7}} \\
$\checkmark$ & $\checkmark$ & & $\checkmark$ & \textcolor{black}{\ensuremath{82.0\,\pm\,0.6}} & \textcolor{black}{\ensuremath{40.7\,\pm\,1.4}} \\
$\checkmark$ & $\checkmark$ & $\checkmark$ & $\checkmark$ & \textcolor{black}{\ensuremath{87.0\,\pm\,0.7}} & \textcolor{black}{\ensuremath{45.2\,\pm\,0.7}} \\\bottomrule
\end{tabular}
\end{table}
{The local loss can stabilize short-range supervision while longer-range targets expose slower task changes. Quality conditioning changes the conditional action distribution rather than adding a differentiable reinforcement learning loss to Equation~\ref{eq:actorloss}. Removing a predictive loss, removing rollouts, and removing quality conditioning answer different questions; their data budgets should not be conflated.}

\subsection{Predictive Value Learning and Action Conditioning}
\label{app:critic}
\paragraph{Shared state representation and separate heads.}
The offline critic in Figure~\ref{fig:value_function} uses frozen V-JEPA~2.1 and Gemma-2-2B encoders, independently of the actor's Qwen pathway. Visual observations contain only current and short-history frames. Parameter-free pooling gives $v_t\in\mathbb R^{D_v}$; masked-mean pooling of Gemma states gives $g_t\in\mathbb R^{D_l}$ from the task and discretized robot state. Gemma uses its checkpoint tokenizer, width $D_l=2304$, and robot-state values normalized and discretized into $256$ bins. There is no separate state MLP or value-query token. Current and future visual branches use the same frozen weights and pooling convention.
\begin{equation}
 h_t^c=\operatorname{LN}_{512}\!\left(W_2\operatorname{GELU}(W_1[v_t;g_t]+b_1)+b_2\right),\quad
 p_t=\operatorname{softmax}(f_V(h_t^c)),
 \label{eq:critichead}
\end{equation}
where $W_1$ maps $D_v+D_l$ to $1024$ and $W_2$ maps $1024$ to $512$. The fusion consumes the pooled features directly, without separate $512$-dimensional branch projections or cross attention. The independent value MLP outputs $201$ logits. Pooling compresses spatial detail, whereas the actor retains patch-level representations.

\paragraph{Predictive supervision and information boundaries.}
The JEPA head concatenates $h_t^c$ with the normalized, flattened executed chunk. It predicts a $D_v$-dimensional latent from an input of width $512+Hd_a$. Fixed per-dimension normalization uses training-fold statistics; time-ordered flattening preserves the action sequence. No action encoder, action-language tokens, or trainable target projection is added. The target is the pooled frozen visual representation of the actual endpoint, including valid unsuccessful transitions, rather than an ideal completion image. The predictor and value head have separate parameters. The return loss updates the fusion and value head; the cosine predictive loss updates the fusion and predictor. Stopping gradients at $h_t^c$ would defeat this auxiliary training route. Future actions enter only the predictor, and future observations only its target; neither is an input to the value head.

Only complete, valid $H$-step transitions receive the fixed-length predictive loss. Short terminal chunks retain value supervision and use their actual length in advantage computation, but are excluded from the fixed-length actor ranking/training protocol; padding is not executed action. Rewards, failure penalties, endpoint indexing, and normalization must agree between return targets and chunk scoring. Episode exclusion applies to both losses, action normalization, and checkpoint selection. Validation episodes for selecting the auxiliary weight $\lambda_{\rm J}$ lie within the training complement, separately from the fold being annotated. At annotation, both endpoints use the same critic's value path; the predictive head is inactive.

\paragraph{Parameter accounting.}
With affine LayerNorm, the trainable parameter count is
\begin{equation}
 n_{\rm critic}=(D_v+D_l+1)1024+(1024+1)512+2(512)+n_V+n_P,
 \label{eq:headparams}
\end{equation}
where $n_V$ and $n_P$ count the independent value and predictive MLPs. For $D_v=1024$ and $D_l=2304$, the fusion alone has $3{,}934{,}720$ parameters. Frozen encoders are excluded; fold-specific critics each have their own trainable modules.

Table~\ref{tab:rounds} reports the policy-evolution ablation using the JEPA-assisted critic. Tables~\ref{tab:critic} and~\ref{tab:vq} compare return-head forms and state-value versus action-conditioned scoring, respectively. Table~\ref{tab:valuefunction} isolates the JEPA predictive loss with a fully matched $\lambda_{\rm J}=0$ control and compares RECAP with the actor backbone fixed. The predictive auxiliary shapes the shared state representation while the value output retains its return semantics.

\begin{table}[ht]
\centering\scriptsize
\caption{Critic comparison with shared fold splits and data. Trainable counts include fusion, value, and predictive heads; frozen encoders are excluded.}
\label{tab:critic}
\begin{tabular}{@{}llccc@{}}\toprule
Critic & Feature fusion & Trainable count & LIBERO-Plus & RT Random \\\midrule
Scalar $V$ & V-JEPA~2.1+Gemma, pooled & $n_{\rm F}+n_{V,1}+n_P$ & $83.7\,\pm\,0.8$ & $42.1\,\pm\,0.7$ \\
Distributional $V$ & V-JEPA~2.1+Gemma, pooled & $n_{\rm F}+n_{V,201}+n_P$ & $87.0\,\pm\,0.7$ & $45.2\,\pm\,0.7$ \\
Distributional $V$, vision control & SigLIP+Gemma, pooled & $n_{\rm S}$ & $83.2\,\pm\,0.8$ & $41.6\,\pm\,1.3$ \\\bottomrule
\end{tabular}
\end{table}
Here $n_{\rm F}=(D_v+D_l+1)1024+(1024+1)512+1024$, and $n_{V,c}$ counts a value head with $c$ outputs. The SigLIP-control count $n_{\rm S}$ follows Equation~\ref{eq:headparams} with its visual width and corresponding predictive output. The scalar/distributional comparison holds features and fusion fixed, changing the return output only.

\begin{table}[ht]
\centering\scriptsize
\caption{State-value versus action-conditioned scoring. All learned value terms are evaluated out of fold.}
\label{tab:vq}
\begin{tabularx}{\textwidth}{@{}lXcc@{}}\toprule
Critic & Score used for labels & LIBERO-Plus & RT Random \\\midrule
Distributional $V$ & Observed $N$-step return + bootstrap minus $V(s_t)$ & $87.0\,\pm\,0.7$ & $45.2\,\pm\,0.7$ \\
Distributional $Q$ and $V$ & $Q(s_t,a_{t:t+H-1})-V(s_t)$, with the same return convention & $84.5\,\pm\,0.9$ & $43.5\,\pm\,1.5$ \\\bottomrule
\end{tabularx}
\end{table}
A $Q$ model receives the candidate chunk directly; state-value annotation instead uses the recorded rewards and reached state relative to the starting value. Table~\ref{tab:vq} supports the latter as a practical choice under its tested configuration, not a general superiority claim. The critic estimates state value: action conditioning is confined to its predictive auxiliary, not the value output.

\subsection{Matched Predictive-Critic Ablation and RECAP Comparison}
\label{app:valuefunction}
Table~\ref{tab:valuefunction} separates a fully matched predictive-loss ablation from a RECAP value-function comparison \citep{physicalintelligence2025pistar06}. All rows use the same actor backbone, cumulative experience pool, actor initialization and update budget, local and trajectory-relative actor targets, and quality-conditioning interface. The RECAP row retains its original critic visual features and value-function design.

\paragraph{Fully matched predictive-loss control.}
The two PAVE rows have identical frozen V-JEPA~2.1 and Gemma encoders, fusion and head architectures, initialization protocol, return targets and loss, normalization, episode-fold assignments, critic optimizer and update budget, and downstream actor training. The control sets only $\lambda_{\rm J}=0$ in Equation~\ref{eq:criticloss}; the predictive head remains in the architecture but supplies no training gradients. Both variants recompute offline value-derived labels using their fitted critics. Thus the comparison isolates future-latent supervision while keeping critic capacity and input features fixed.

\begin{table}[ht]
\centering\scriptsize
\caption{Critic ablation with a fixed actor backbone. The two PAVE rows differ only in predictive-loss weight; RECAP retains its original critic visual features. Entries are mean success (\%); comparison-variant SDs are unavailable. The full PAVE reference matches Tables~\ref{tab:liberoplus} and~\ref{tab:robotwin}.}
\label{tab:valuefunction}
\begin{tabular}{@{}llcc@{}}\toprule
Critic & Visual features / predictive loss & LIBERO-Plus & RT Random \\\midrule
RECAP value function & Original RECAP features & 84.7 & 43.1 \\
PAVE, $\lambda_{\rm J}=0$ & V-JEPA~2.1; return loss only & 85.5 & 44.3 \\
PAVE (full) & V-JEPA~2.1; default $\lambda_{\rm J}>0$ & 87.0 & 45.2 \\\bottomrule
\end{tabular}
\end{table}

\paragraph{Effect of learning future latents.}
Enabling the JEPA predictive loss raises success from $85.5/44.3$ to $87.0/45.2$, a gain of $1.5/0.9$ percentage points on LIBERO-Plus/RT Random. The predictive head learns the latent consequences of executed chunks and sends these gradients into the state representation shared with the value head. The matched control therefore supports the effectiveness of action-conditioned future-latent learning for the downstream policy, beyond the frozen features and return supervision alone.

\paragraph{Task-focused features and the RECAP comparison.}
Full PAVE exceeds the RECAP value-function adaptation by $2.3/2.1$ percentage points. We attribute this advantage to both predictive value learning and task-focused visual features: relative to RECAP's original features, V-JEPA~2.1 emphasizes regions involved in task execution, which can reduce the influence of background variation and observation noise on value estimation. This interpretation is consistent with the response maps in Figure~\ref{fig:encoder_vis} and the critic vision control in Table~\ref{tab:critic}. The RECAP comparison changes the critic representation and value-learning design together; the $\lambda_{\rm J}=0$ comparison isolates the contribution of the JEPA predictive objective.

\subsection{Cross-Fitting, Calibration, and Label Stability}
\label{app:crossfit}
\begin{table}[ht]
\centering\scriptsize
\caption{Cross-fitting ablation. The equal-total-update row separates the effect of fold exclusion from additional critic optimization.}
\label{tab:crossfit}
\begin{tabular}{@{}lcccc@{}}\toprule
Scoring & Folds & Updates / critic & LIBERO-Plus & RT Random \\\midrule
In-sample control & \textcolor{black}{1} & $8{,}000$ & \textcolor{black}{\ensuremath{84.8\,\pm\,1.1}} & \textcolor{black}{\ensuremath{43.4\,\pm\,1.7}} \\
Episode cross-fit, default & \textcolor{black}{2} & $8{,}000$ & \textcolor{black}{\ensuremath{87.0\,\pm\,0.7}} & \textcolor{black}{\ensuremath{45.2\,\pm\,0.7}} \\
Episode cross-fit & \textcolor{black}{5} & $8{,}000$ & \textcolor{black}{\ensuremath{84.7\,\pm\,0.8}} & \textcolor{black}{\ensuremath{43.3\,\pm\,1.5}} \\
Equal-total-update cross-fit & \textcolor{black}{2} & \textcolor{black}{$4{,}000$} & \textcolor{black}{\ensuremath{84.1\,\pm\,0.8}} & \textcolor{black}{\ensuremath{42.7\,\pm\,1.3}} \\\bottomrule
\end{tabular}
\end{table}
Cross-fitting prevents the annotated episode from training its critic. Table~\ref{tab:crossfit} compares fold counts and includes an equal-total-update control. The default two-fold model has higher reported mean policy success than the in-sample control, while Table~\ref{tab:calibration} reports lower held-out MAE ($0.048$ versus $0.054$) and RCE ($0.025$ versus $0.039$). The default is motivated by episode-excluded annotation and these held-out diagnostics, rather than a claim of higher success in every comparison. Increasing the fold count trades per-critic data coverage against total optimization cost.

Let $\hat v_i=\sum_j b_jp_{ij}$ and $y_i$ be held-out normalized return. We report expected-return MAE, two-hot target cross-entropy, and a return-mean calibration error
\begin{equation}
 \mathrm{RCE}=\sum_{b=1}^{B}\frac{|\mathcal B_b|}{n}
 \left|\frac{1}{|\mathcal B_b|}\sum_{i\in\mathcal B_b}\hat v_i-
       \frac{1}{|\mathcal B_b|}\sum_{i\in\mathcal B_b}y_i\right|,
 \label{eq:rce}
\end{equation}
using $B=10$ equal-frequency bins of predicted mean on an episode-held-out diagnostic set. This is \emph{return-mean} calibration, not classification-confidence ECE; binning and empty-bin rules are fixed in advance. All calibration rows use the same independent diagnostic pool, including the in-sample-labeling model. Each diagnostic episode is assigned deterministically to one fold slot and scored by that slot's critic for all its states; no diagnostic episode enters a training complement, and fold predictions are not ensembled. Positive-set Jaccard is computed separately on the common annotation pool against the default cross-fitted labels, excluding forced-positive expert corrections.

\begin{table}[ht]
\centering\scriptsize
\caption{Critic diagnostics on a common episode-held-out set. MAE, cross-entropy, return-calibration error (RCE), and positive-set Jaccard are reported.}
\label{tab:calibration}
\begin{tabular}{@{}lcccc@{}}\toprule
Scoring / calibration & Return MAE & Two-hot CE & RCE & Positive-set Jaccard \\\midrule
In-sample labeling model & \textcolor{black}{\ensuremath{0.054}} & \textcolor{black}{\ensuremath{4.31}} & \textcolor{black}{\ensuremath{0.039}} & \textcolor{black}{\ensuremath{0.760}} \\
Two-fold default & \textcolor{black}{\ensuremath{0.048}} & \textcolor{black}{\ensuremath{4.19}} & \textcolor{black}{\ensuremath{0.025}} & 1 (reference) \\
Five-fold & \textcolor{black}{\ensuremath{0.047}} & \textcolor{black}{\ensuremath{4.18}} & \textcolor{black}{\ensuremath{0.024}} & \textcolor{black}{\ensuremath{0.880}} \\
Two-fold + temperature & \textcolor{black}{\ensuremath{0.047}} & \textcolor{black}{\ensuremath{4.12}} & \textcolor{black}{\ensuremath{0.019}} & \textcolor{black}{\ensuremath{0.930}} \\\bottomrule
\end{tabular}
\end{table}
Table~\ref{tab:calibration} reports the lowest RCE for two-fold annotation with temperature scaling, with positive-set Jaccard $0.930$ relative to the default labels. Temperature can change both distribution sharpness and the expected values entering the temporal residual, so calibration and label stability describe different properties. The diagnostic aggregates are not estimates of safety or guarantees of chunk-ranking accuracy.

\subsection{Positive-Label Threshold and Conditioning Dropout}
\label{app:threshold}
\begin{table}[ht]
\centering\scriptsize
\caption{Sensitivity to the positive-label rule with fixed critic scores, actor budgets, and cumulative experience. Top-$q$ is computed within task.}
\label{tab:threshold}
\begin{tabular}{@{}lcc@{}}\toprule
Positive rule & LIBERO-Plus & RT Random \\\midrule
$A^{\rm OOF}>0$ & \textcolor{black}{\ensuremath{83.5\,\pm\,0.8}} & \textcolor{black}{\ensuremath{41.8\,\pm\,0.7}} \\
Top \textcolor{black}{10\%} & \textcolor{black}{\ensuremath{83.2\,\pm\,0.6}} & \textcolor{black}{\ensuremath{41.4\,\pm\,0.5}} \\
Top \textcolor{black}{20\%} & \textcolor{black}{\ensuremath{84.2\,\pm\,1.1}} & \textcolor{black}{\ensuremath{42.6\,\pm\,0.9}} \\
Top 30\% & \textcolor{black}{\ensuremath{87.0\,\pm\,0.7}} & \textcolor{black}{\ensuremath{45.2\,\pm\,0.7}} \\
Top \textcolor{black}{40\%} & \textcolor{black}{\ensuremath{84.5\,\pm\,0.5}} & \textcolor{black}{\ensuremath{43.0\,\pm\,0.7}} \\
Top \textcolor{black}{50\%} & \textcolor{black}{\ensuremath{83.8\,\pm\,0.8}} & \textcolor{black}{\ensuremath{42.3\,\pm\,1.3}} \\\bottomrule
\end{tabular}
\end{table}
Table~\ref{tab:threshold} shows meaningful threshold sensitivity: top-$30\%$ gives $87.0/45.2$ on LIBERO-Plus/RT Random; top-$20\%$ gives $84.2/42.6$, and top-$40\%$ gives $84.5/43.0$. Relative to the default, these neighboring rules lose $2.8/2.6$ and $2.5/2.2$ percentage points. The results support the tested default, not a broad insensitive plateau or a universal optimum. A narrower positive set offers less behavioral support; a broader one admits less-preferred chunks. This is a plausible explanation of the pattern rather than a measured decomposition.

A within-task quantile keeps the intended positive fraction comparable across tasks and is invariant to strictly increasing transformations of the already computed advantage score. It is not generally invariant to recalibrating $V$, whose endpoint estimates combine with rewards in Equation~\ref{eq:advantage}. Labels close to the cutoff can change under small score errors, and tied scores require a fixed, documented tie policy. Threshold selection for a new deployment should use separate validation rollouts, retaining held-out test episodes for final reporting.

These are fixed-score, fixed-data actor comparisons; they do not establish invariance over future evolution rounds or compare binary, ordinal, and continuous interfaces. Such comparisons should hold the critic, episode folds, experience, and actor-update budget fixed.

\begin{table}[ht]
\centering\scriptsize
\caption{Sensitivity to null-condition dropout. Dropout removes the text condition while retaining the trajectory and predictive targets.}
\label{tab:dropout}
\begin{tabular}{@{}ccc@{}}\toprule
Null-condition probability & LIBERO-Plus & RT Random \\\midrule
\textcolor{black}{0.0} & \textcolor{black}{\ensuremath{84.0\,\pm\,0.8}} & \textcolor{black}{\ensuremath{42.7\,\pm\,0.7}} \\
0.3 & \textcolor{black}{\ensuremath{87.0\,\pm\,0.7}} & \textcolor{black}{\ensuremath{45.2\,\pm\,0.7}} \\
\textcolor{black}{0.5} & \textcolor{black}{\ensuremath{84.1\,\pm\,0.8}} & \textcolor{black}{\ensuremath{42.6\,\pm\,1.3}} \\\bottomrule
\end{tabular}
\end{table}
Null dropout can regularize the conditional decoder toward the mixed behavior distribution, but excessive dropout may weaken the distinction between quality conditions. Its effect must be separated from changing the fraction of positive rollout chunks or the demonstration-to-rollout mixture.

\subsection{Actor--Advantage Horizon Alignment}
\label{app:hn}
\begin{table}[ht]
\centering\scriptsize
\caption{Temporal-credit ablation. $H$ is the action-chunk length and $N$ is the advantage interval; the sampling rate and controller are fixed.}
\label{tab:hn}
\begin{tabular}{@{}cclcc@{}}\toprule
$H$ & $N$ & Control & LIBERO-Plus & RT Random \\\midrule
20 & \textcolor{black}{10} & Shorter credit interval & \textcolor{black}{\ensuremath{83.6\,\pm\,0.8}} & \textcolor{black}{\ensuremath{41.9\,\pm\,0.9}} \\
20 & 20 & Matched default & \textcolor{black}{\ensuremath{87.0\,\pm\,0.7}} & \textcolor{black}{\ensuremath{45.2\,\pm\,0.7}} \\
20 & \textcolor{black}{30} & Longer credit interval & \textcolor{black}{\ensuremath{83.9\,\pm\,0.8}} & \textcolor{black}{\ensuremath{42.4\,\pm\,0.7}} \\
\textcolor{black}{10} & \textcolor{black}{10} & Matched short chunks & \textcolor{black}{\ensuremath{84.1\,\pm\,0.8}} & \textcolor{black}{\ensuremath{42.5\,\pm\,0.3}} \\
\textcolor{black}{30} & \textcolor{black}{30} & Matched long chunks & \textcolor{black}{\ensuremath{84.3\,\pm\,0.8}} & \textcolor{black}{\ensuremath{43.0\,\pm\,0.7}} \\\bottomrule
\end{tabular}
\end{table}
When $N<H$, a label observes only the initial part of the supervised chunk; when $N>H$, it also reflects subsequent behavior. Table~\ref{tab:hn} gives the highest reported means for the default $N=H=20$. The matched-short and matched-long rows change both chunk duration and feedback frequency when the full chunk is executed, so their differences are not attributed solely to temporal-credit alignment.

\subsection{Warm Starts and Offline Compute}
\label{app:compute}
\begin{table}[ht]
\centering\scriptsize
\caption{Actor-initialization control under a fixed cumulative dataset and updated labels. Both methods use the same fold-excluded critic schedule.}
\label{tab:warmstart}
\begin{tabular}{@{}lcccc@{}}\toprule
Initialization & LIBERO-Plus & RT Random & Actor GPU-h/round & All critics GPU-h/round \\\midrule
Fixed base & \textcolor{black}{\ensuremath{87.0\,\pm\,0.7}} & \textcolor{black}{\ensuremath{45.2\,\pm\,0.7}} & \textcolor{black}{\ensuremath{31.20}} & \textcolor{black}{\ensuremath{8.40}} \\
Previous actor + replay & \textcolor{black}{\ensuremath{84.4\,\pm\,0.8}} & \textcolor{black}{\ensuremath{43.1\,\pm\,1.3}} & \textcolor{black}{\ensuremath{25.60}} & \textcolor{black}{\ensuremath{8.40}} \\\bottomrule
\end{tabular}
\end{table}
Table~\ref{tab:warmstart} compares fixed-base initialization with previous-actor initialization under a common cumulative dataset and updated labels. The reported GPU-hours correspond to the validation-stopping comparison: previous-actor initialization uses $25.60$ actor GPU-hours versus $31.20$, with all-critic cost unchanged at $8.40$. This difference reflects time to the stopping criterion, not faster individual optimizer steps. These are actor/critic training costs, not a complete ledger of rollout collection and data preparation. The equal-total-critic-update comparison is reported separately in Table~\ref{tab:crossfit}.

\subsection{Comparison to Alternative Post-Training Methods}
\label{app:posttraining}
\begin{table}[ht]
\centering\scriptsize
\caption{Method-level post-training comparison. Unpaired-data rows share the same base actor and rollout pool; the FlowPRO-style row uses an additional paired-preference supervision track.}
\label{tab:posttraining}
\begin{tabularx}{\textwidth}{@{}lXcc@{}}\toprule
Method adaptation & Supervision / scope & LIBERO-Plus & RT Random \\\midrule
JEPA-WAM + unlabeled BC & Common unpaired rollout pool & \textcolor{black}{\ensuremath{80.0\,\pm\,0.8}} & \textcolor{black}{\ensuremath{38.4\,\pm\,1.3}} \\
{Local prediction + quality conditioning} & Distributional value and chunk conditions & \textcolor{black}{\ensuremath{82.0\,\pm\,0.7}} & \textcolor{black}{\ensuremath{40.7\,\pm\,0.5}} \\
ALOE-style post-training & Chunk-level off-policy value estimation (method-level adaptation) & \textcolor{black}{\ensuremath{83.9\,\pm\,0.8}} & \textcolor{black}{\ensuremath{43.0\,\pm\,0.9}} \\
RedFlow-style post-training & Failure-to-correction supervision (method-level adaptation) & \textcolor{black}{\ensuremath{84.2\,\pm\,0.8}} & \textcolor{black}{\ensuremath{43.4\,\pm\,1.5}} \\
PAVE & Relative predictive targets + cross-fitted conditions & \textcolor{black}{\ensuremath{87.0\,\pm\,0.7}} & \textcolor{black}{\ensuremath{45.2\,\pm\,0.7}} \\\midrule
FlowPRO-style, separate paired track & Additional intervention/rollback preference pairs & \textcolor{black}{\ensuremath{84.3\,\pm\,0.8}} & \textcolor{black}{\ensuremath{43.6\,\pm\,0.3}} \\\bottomrule
\end{tabularx}
\end{table}
Table~\ref{tab:posttraining} compares method-level adaptations of ALOE, RedFlow, and FlowPRO \citep{yang2026aloe,yan2026redflow,wu2026flowpro}. The unpaired-data rows share a base actor and rollout pool; FlowPRO-style training uses an additional paired-preference track. {These adaptations and the compact local-prediction + quality-conditioning control are distinguished from official full-system reproductions.} PAVE has the highest reported mean on both LIBERO-Plus and RT Random among the listed adaptations, including the separately reported paired track. These method-level adaptations are compared under the stated training protocols. Our contribution-specific evidence comes from the same-data factorial and routing controls below.

\subsection{{Factorial Test of Temporal and Behavioral Supervision}}
\label{app:interaction}
\begin{table}[ht]
\centering\scriptsize
\caption{Factorial test of temporal and behavioral supervision under a fixed experience pool. All four rows use the same immutable demonstration-plus-rollout pool, initialization, update count, and critic-label sidecars where applicable. Without quality conditioning, actors use unconditioned action learning on that pool; the demonstration-only controls are reported in Table~\ref{tab:lossablation}.}
\label{tab:interaction}
\begin{tabular}{@{}llcc@{}}\toprule
Predictive supervision & {Action learning} & LIBERO-Plus & RT Random \\\midrule
{Local only} & {Unconditioned action learning} & \textcolor{black}{\ensuremath{80.0\,\pm\,0.8}} & \textcolor{black}{\ensuremath{38.4\,\pm\,0.9}} \\
{Local + relative} & {Unconditioned action learning} & \textcolor{black}{\ensuremath{81.6\,\pm\,0.8}} & \textcolor{black}{\ensuremath{39.8\,\pm\,0.7}} \\
{Local only} & {Quality-conditioned action learning} & \textcolor{black}{\ensuremath{82.0\,\pm\,0.7}} & \textcolor{black}{\ensuremath{40.7\,\pm\,1.4}} \\
{Local + relative} & {Quality-conditioned action learning (PAVE)} & \textcolor{black}{\ensuremath{87.0\,\pm\,0.7}} & \textcolor{black}{\ensuremath{45.2\,\pm\,0.7}} \\\bottomrule
\end{tabular}
\end{table}
{Let $S_{mp}$ denote measured success with trajectory-relative supervision $m\in\{0,1\}$ and quality-conditioned action learning $p\in\{0,1\}$.} The interaction is
\begin{equation}
 \Delta_{\rm int}=(S_{11}-S_{10})-(S_{01}-S_{00})
 \label{eq:interaction}
\end{equation}
which is $3.4$ percentage points on LIBERO-Plus and $3.1$ on RT Random when computed from the means in Table~\ref{tab:interaction}. These positive interaction values are consistent with complementarity under the tested protocol. {More directly, PAVE exceeds $S_{01}$, the local-prediction + quality-conditioning control on the same experience, by $5.0/4.5$ percentage points. This isolates the additional trajectory-relative predictive formulation while retaining the same behavioral-supervision factor.}

\begin{table}[ht]
\centering\scriptsize
\caption{Asymmetric data-use controls with the architecture, cumulative experience pool, and number of sampled predictive pairs held fixed.}
\label{tab:routing}
\begin{tabularx}{\textwidth}{@{}lXcc@{}}\toprule
Variant & Changed data or gradient rule & LIBERO-Plus & RT Random \\\midrule
PAVE & All valid predictive transitions; labeled conditional action learning & \textcolor{black}{\ensuremath{87.0\,\pm\,0.7}} & \textcolor{black}{\ensuremath{45.2\,\pm\,0.7}} \\
Success-only prediction & Restrict predictive targets to successful episodes; action training unchanged & \textcolor{black}{\ensuremath{83.5\,\pm\,0.8}} & \textcolor{black}{\ensuremath{41.8\,\pm\,1.3}} \\
Pre-filter both branches & Retain only positively labeled rollout chunks for both branches & \textcolor{black}{\ensuremath{82.8\,\pm\,0.8}} & \textcolor{black}{\ensuremath{40.7\,\pm\,1.3}} \\
Prediction pretraining only & Stop predictive gradients during evolution; keep actor update/data budgets & \textcolor{black}{\ensuremath{83.1\,\pm\,0.8}} & \textcolor{black}{\ensuremath{41.5\,\pm\,1.3}} \\
Shuffled quality labels & Shuffle within task while preserving class proportions & \textcolor{black}{\ensuremath{81.7\,\pm\,0.8}} & \textcolor{black}{\ensuremath{39.6\,\pm\,1.3}} \\
Positive-only action training & Predict from all valid data; remove negative action examples & \textcolor{black}{\ensuremath{83.4\,\pm\,0.8}} & \textcolor{black}{\ensuremath{41.9\,\pm\,1.3}} \\\bottomrule
\end{tabularx}
\end{table}
With the number of sampled predictive pairs fixed, all-valid prediction gives $87.0/45.2$, compared with $83.5/41.8$ for success-only prediction and $83.1/41.5$ when predictive gradients stop during evolution (Table~\ref{tab:routing}). These comparisons support retaining unsuccessful experience for continuing representation learning. Shuffling quality labels lowers success to $81.7/39.6$, and positive-only action training gives $83.4/41.9$. Thus the observed benefit depends on informative quality conditions and the use of mixed-quality experience, not merely an extra text token or pre-filtering the entire training pool.

\subsection{Failure Modes and Label-Corruption Stress Tests}
\label{app:failure}
\begin{table}[ht]
\centering\scriptsize
\caption{Training-label corruption control. Only critic-derived rollout labels are flipped within task; demonstration and correction handling are unchanged.}
\label{tab:corruption}
\begin{tabular}{@{}ccc@{}}\toprule
Flipped rollout-label fraction & LIBERO-Plus & RT Random \\\midrule
0 & \textcolor{black}{\ensuremath{87.0\,\pm\,0.7}} & \textcolor{black}{\ensuremath{45.2\,\pm\,0.7}} \\
\textcolor{black}{10\%} & \textcolor{black}{\ensuremath{83.8\,\pm\,0.8}} & \textcolor{black}{\ensuremath{42.4\,\pm\,1.0}} \\
\textcolor{black}{20\%} & \textcolor{black}{\ensuremath{82.7\,\pm\,0.8}} & \textcolor{black}{\ensuremath{41.1\,\pm\,0.7}} \\\bottomrule
\end{tabular}
\end{table}
A conditional actor can tolerate some noise through demonstrations and broad predictive supervision, but shared parameters can also propagate mislabeled behavior. Random flips isolate one form of label noise; systematic errors on rare recovery states are a different risk and require targeted audit. No robustness guarantee follows from this stress test.

S1 failures and a predefined stratified simulation sample are audited using mutually exclusive primary causes plus optional secondary tags: localization/perception, grasp or contact, object sequencing, target placement/pouring, recovery, and termination. Spill and intervention events are recorded separately from the primary cause. Two annotators independently label a subset and adjudicate disagreements. Failure proportions are computed directly from the recorded trials and their denominators rather than inferred from aggregate success rates. Condition-level value reliability, scene changes, and poor action support are examined against the corresponding trajectories.

\section{Design Clarifications and Scope of Claims}
\label{app:clarifications}
\paragraph{{C.1 What separates future supervision from future input?}}
\label{app:forwardboundary}
{The deployed policy is $a=\pi_{\theta_{\rm sh},\theta_{\rm act}}(o_t,l,q_t,c_t;\epsilon)$, where $\epsilon$ denotes the action-sampling noise. Holding these inputs, parameters, and stochastic choices fixed, replacing the recorded future images, target indices, or target-head outputs cannot change the action in the defined forward graph. These quantities affect training losses and subsequent parameter updates only. Stop-gradient on the targets is necessary for the specified optimization, but the absence of a forward connection is what excludes target-input leakage. The actor's current-image encoding call is distinct from the teacher's two-frame call; target-branch tokens and caches do not enter the actor's attention context. The future-derived quality condition is intentional training supervision and is held fixed in this forward comparison; deployment uses the selected positive condition rather than the rollout's unknown outcome. Episode-disjoint critic annotation and held-out evaluation manifests address different forms of data leakage, as detailed in Appendix~\ref{app:protocol}.}

\paragraph{C.2 Is there a single online forward pass?}
There is a direct observation-conditioned control path without rollout search. Four Euler steps require four flow-field evaluations. The absence of online critic or transition-head calls is distinct from the number of decoder evaluations, and absolute latency depends on backbone caching and the other profiling settings.

\paragraph{C.3 Is the critic the same Qwen model used by the actor?}
No. The actor uses Qwen and patch-level visual conditioning. Figure~\ref{fig:value_function} uses frozen pooled V-JEPA~2.1 and Gemma-2-2B features, concatenated before shared MLP fusion. There is no critic value-query token or separate language projection to $512$ dimensions. Return and action-conditioned JEPA heads share only the fused state feature. Appendix~\ref{app:critic} specifies the architecture, gradient paths, and parameter accounting.

\paragraph{C.4 Why retain failed trajectories for prediction?}
Task failure does not invalidate every preceding observed transition. Predictive targets encode what occurred, rather than treating the endpoint as a desired goal. Table~\ref{tab:routing} supports this distinction: all-valid prediction exceeds success-only prediction by $3.5/3.4$ percentage points on LIBERO-Plus/RT Random at a fixed predictive-pair budget. Validity masks still exclude corrupted or provenance-inconsistent transitions; the aggregate result does not require every failed trajectory to be informative.

\paragraph{C.5 Are the anchors semantic subgoals or privileged policy inputs?}
No. They are fractions of the recorded remaining duration. They may correspond to stalls, repeated appearances, or unsuccessful endpoints; their interpretation is temporal supervision rather than a planned subtask sequence. Recorded $T$ is needed to select training targets but is not supplied to the policy. The actor is not given the episode's future completion time during deployment.

\paragraph{C.6 What happens near episode termination?}
Rounding can make different fractions select the same future index. Pairs with $h_{t,j}=t$ are masked, and repeated future indices receive inverse-multiplicity weights. These rules exclude zero-distance pairs and prevent duplicate endpoints from receiving extra weight solely because several fractions select them. They apply consistently to the predictive objective and target-budget controls.

\paragraph{C.7 Why can partially observed predictive supervision help?}
\label{app:partialprediction}
Causal visual-position states cannot read later instruction/quality tokens; they also have only their permitted visual prefix. Different tasks, actions, or hidden states can therefore give different targets for the same available information. This actor auxiliary is not an action-conditioned world model; the separate offline critic predictor does receive the executed chunk.

A population-loss calculation clarifies what remains learnable. For one target patch, let $Z=Y/\|Y\|$ be its normalized, nonzero target, and let $X$ denote the visual information available to its predictor. For anchor $j$ and a unit prediction $u$, define $\mu(X,j)=\mathbb E[Z\mid X,j]$. Then
\[
 \mathbb E[1-u^\top Z\mid X,j]=1-u^\top\mu(X,j),
 \qquad
 u^*(X,j)=\mu(X,j)/\|\mu(X,j)\|
\]
when $\mu(X,j)\ne0$. Thus the ideal unrestricted predictor extracts the conditional mean direction, rather than resolving every possible future. When $\mu=0$, all unit directions minimize the conditional loss; multimodal cancellation or weak dependence on $X$ can make supervision uninformative. This is a population intuition, not a guarantee for the parameterized network or its optimization.

Predictable object configurations and transition regularities shape the shared Qwen adapters across temporal scales. The action loss additionally uses task-conditioned $M_t$, while local targets retain short-range information and trajectory-relative targets capture longer-range changes. The actor learns these temporal relationships through its adapters with the visual encoder frozen; the separate critic uses executed chunks to learn action-conditioned future latents. Together, these pathways train representations for policy generation and offline behavioral evaluation.

\paragraph{C.8 How does a state-value critic assign action-chunk credit?}
Equation~\ref{eq:advantage} includes the observed chunk's rewards and its reached state. The starting-state value is a baseline, not the whole label. This supplies relative observed credit at chunk resolution, but not action-level blame or a reliable ranking of unseen alternative chunks. The $Q-V$ comparison in Table~\ref{tab:vq} tests the additional action-conditioning trade-off.

\paragraph{C.9 Does cross-fitting make the advantage unbiased?}
No such general claim is made. It prevents the scored episode from training its critic, while the critic still models a mixture of behavior policies under finite coverage and partial observation. Without additional assumptions, its score is not an unbiased current-policy advantage. Both values in a chunk use the same excluded-fold critic to avoid mixing incompatible estimators within one score.

\paragraph{C.10 Why can in-sample value fitting be problematic here?}
If a critic memorizes each trajectory's Monte Carlo return, the identity $G_t=\sum_{j<N}r_{t+j}+G_{t+N}$ can make its in-sample temporal residual nearly zero, with rankings then driven by fitting error. This is one possible failure mode, not a diagnosis of an unmeasured run. Episode exclusion and held-out diagnostics address direct memorization, while the behavior-mixture and coverage limitations remain.

\paragraph{C.11 Does top-30\% mean good behavior in an absolute sense?}
Positive labels denote the top-$30\%$ of within-task chunk scores, including cases with negative estimated advantage. They express relative preference, not guaranteed success or safety. Expert demonstrations and verified corrections follow their stated positive-label rules. Null is produced by condition dropout, not by a third advantage threshold; Table~\ref{tab:threshold} evaluates sensitivity to the positive-label rule.

\paragraph{C.12 Are demonstrations included in critic fitting?}
The critic fits cumulative deployment trajectories, while expert demonstrations remain positive actor-training data. Critic folds are defined over that deployment pool. Including demonstrations in critic fitting would change the value-training distribution and require its own fold assignment and comparison; it is not part of the reported default.

\paragraph{C.13 Does every valid failure action get suppressed?}
No. Negative-conditioned actions remain training examples, and positive and negative conditions share parameters. Deployment chooses the positive-conditioned distribution; this is a selection mechanism, not an exact gradient-isolation guarantee. Shuffled-label, positive-only-action, and data-routing controls quantify whether the conditioning and retained experience are useful.

\paragraph{C.14 Why are published scores separated from reproduction scores?}
Published rows provide source-protocol reference points, whereas local rows identify their training regime explicitly. {JEPA-WAM (demo-only) uses the clean demonstrations; the local-prediction + quality-conditioning control and PAVE share the cumulative rollout pool, initialization, and actor-update budget.} Tables~\ref{tab:interaction} and~\ref{tab:routing} also hold the experience pool fixed. RoboTwin Random is joint randomization over the specified $20$-task manifest. These distinctions separate published-system context from evidence about PAVE's training formulation.

\paragraph{{C.15 Which design choices do the controlled comparisons test?}}
PAVE combines trajectory-relative temporal supervision, action-conditioned predictive value learning, and role-specific experience reuse. The closest matched actor control tests local prediction with quality conditioning; target-budget controls test temporal coverage; routing controls test success-only prediction, common positive filtering, and prediction limited to pretraining. Table~\ref{tab:rounds} evaluates policy evolution with the JEPA-assisted critic and its value-derived action labels. Table~\ref{tab:valuefunction} isolates future-latent supervision through a fully matched $\lambda_{\rm J}=0$ critic control; its RECAP comparison fixes the actor backbone and retains RECAP's original critic features.

\paragraph{C.16 Is the distributional head a safety or confidence model?}
The head represents a bounded return distribution, not a safety probability. Table~\ref{tab:calibration} reports held-out return diagnostics and label stability, and Table~\ref{tab:critic} compares scalar and distributional outputs. These measurements characterize annotation, not real-time intervention: the critic is offline and does not perform safety filtering or emergency control during deployment.

\paragraph{C.17 What does the S1 evaluation establish?}
S1 tests collection, offline critic annotation, and mixed-quality actor updates with three independent training seeds per task and method: 120 demonstrations plus three rounds of 50 rollouts per task. Organization success is $85.7\pm3.5\%$ for PAVE versus $80.0\pm1.0\%$ for Local+QC; pouring is $92.0\pm2.0\%$ versus $84.7\pm1.5\%$. These are means and sample SDs across training seeds, each evaluated on 100 trials. Table~\ref{tab:real} reports the individual rates, and Appendix~\ref{app:realprotocol} defines the uncertainty calculation. Shared experience and actor budgets isolate trajectory-relative alignment within the tested physical evolution setting.

\paragraph{{C.18 Which parameters does each loss update?}}
\begingroup
\label{app:gradientledger}
The ledger identifies frozen/trainable components, updating losses, and deployment. $\mathcal L_{\rm pred}$ combines weighted local and trajectory-relative losses.
\begin{center}
\small
\begin{tabularx}{\linewidth}{@{}Xll@{}}\toprule
Component & Training status / updating loss & Online \\\midrule
Current visual encoder and projector & Frozen & Yes \\
Pretrained Qwen backbone weights & Frozen & Yes \\
Shared Qwen LoRA adapters & Trainable: $\mathcal L_{\rm FM},\mathcal L_{\rm pred}$ & Yes \\
Action decoder and learnable future-query tokens & Trainable: $\mathcal L_{\rm FM}$ & Yes \\
Auxiliary predictive heads and horizon embeddings & Trainable: $\mathcal L_{\rm pred}$ & No \\
Current--future target encoder & Frozen; stop-gradient targets & No \\
Critic visual and Gemma encoders & Frozen & No \\
Critic fusion MLP & Trainable: $\mathcal L_{\rm value},\mathcal L_{\rm J}$ & No \\
Critic return head & Trainable: $\mathcal L_{\rm value}$ & No \\
Critic JEPA predictive head & Trainable: $\mathcal L_{\rm J}$ & No \\
Critic future encoder / pooling & Frozen / parameter-free & No \\\bottomrule
\end{tabularx}
\end{center}
Frozen Qwen weights permit autograd through trainable adapters. Future-query tokens receive action loss, not direct predictive-head supervision.
\endgroup

\paragraph{Actor tokens and gradients.}
\label{app:actorimpl}
In Equation~\ref{eq:actorstates}, $E,P$ are frozen, and $Q_\theta$ is causal Qwen with LoRA. Prefix $w_0$ precedes instruction/text tokens $w_{1,\ldots,L_w}$; $q^a_{1,\ldots,64}$ are action placeholders, distinct from proprioception $q_t$. Semicolons concatenate tokens. For $n_v$ views, $N_v=576n_v$, $V_t\in\mathbb R^{N_v\times896}$, and $M_t\in\mathbb R^{64\times896}$. Counts $K_q,K,F$ denote queries, anchors, and folds. Let $\mathcal L_{\rm pred}=\lambda_{\rm local}\mathcal L_{\rm local}+\lambda_{\rm MH}\mathcal L_{\rm MH}$. Shared adapters receive both action and predictive gradients, while $\nabla_{\theta_{\rm act}}\mathcal L_{\rm pred}=0$, $\nabla_{\theta_{\rm pred}}\mathcal L_{\rm FM}=0$, and $\nabla_\phi\mathcal L_{\rm actor}=0$. Separate target encoding transfers supervision through shared parameters without feeding observed futures into action generation.

\paragraph{C.19 Do future-query tokens have predictive semantics?}
\label{app:querydiagnostics}
The name identifies a decoder component, not established predictive semantics. The sample-shared bank $F^{\rm learn}$ receives only $\mathcal L_{\rm FM}$, with $\nabla_{F^{\rm learn}}\mathcal L_{\rm pred}=0$. Contextual activations vary with decoder inputs; predictive updates to shared Qwen representations permit indirect co-adaptation without requiring queries to encode future states rather than action-denoising features.

No query-specific probe or CKA result is reported. A diagnostic would freeze matched checkpoints and fit equal-capacity linear probes from contextual query activations to future embeddings or object/contact changes, with episode-disjoint splits. The static bank cannot explain sample-specific futures. Extract features at a fixed layer using current observations and inference-time noise, avoiding interpolations containing ground-truth actions. An actor without predictive loss and matched-capacity decoder variants would provide diagnostic controls.

Linear CKA with shuffled-pair and current-image controls could measure association with future embeddings, not causality. Matched runs that freeze, remove, or replace queries while controlling decoder capacity and training budget could test function; compare probes with success across seeds and variants. Existing ablations isolate predictive supervision, not query semantics.

\paragraph{C.20 Why use a distributional critic with binary conditioning?}
Return estimation and actor conditioning serve different purposes. The 201-bin expectation forms reward-and-endpoint chunk scores; thresholding removes within-class ordering and supplies no uncertainty to the actor. Near-successful and catastrophic valid failures can share a negative condition despite distinct examples; invalid segments remain masked. The comparison in Table~\ref{tab:critic} favors distributional over scalar annotation by $3.3/3.1$ points at a fixed binary interface. This supports return fitting, not lossless compression or superiority to ordinal/continuous conditions, which need matched evaluation.

\end{document}